\documentclass[10pt,twocolumn,letterpaper]{article}

\pdfoutput=1
\usepackage{titling}
\usepackage{iccv}
\usepackage{times}
\usepackage{epsfig}
\usepackage{graphicx}
\usepackage{amsmath}
\usepackage{amssymb}

\usepackage[linesnumbered,ruled,vlined]{algorithm2e} 
\usepackage{amsfonts}
\usepackage{amsthm}
\usepackage{authblk}
\usepackage{booktabs}
\usepackage{bm}
\usepackage{enumitem}
\usepackage{makecell}
\usepackage{multirow}
\usepackage{soul}
\usepackage{subcaption}
\usepackage{tabularx}
\usepackage{xcolor}
\usepackage{xpatch}

\usepackage[breaklinks=true,bookmarks=false]{hyperref}

\iccvfinalcopy 

\def\iccvPaperID{2638} 

\ificcvfinal\fi

\begin{document}

\title{Cap2Det: Learning to Amplify Weak Caption Supervision for Object Detection}

\author[1]{Keren Ye\thanks{Work partially done during an internship at Google}}
\author[1]{Mingda Zhang}
\author[1]{Adriana Kovashka}
\author[2]{Wei Li\thanks{Now at Facebook Inc.}}
\author[2]{Danfeng Qin}
\author[2]{Jesse Berent}
\affil[1]{Department of Computer Science, University of Pittsburgh, Pittsburgh PA, USA}
\affil[2]{Google Research, Zurich, Switzerland}
\affil[ ]{\tt\small{\{yekeren, mzhang, kovashka\}@cs.pitt.edu \hspace{0.4cm} lwthucs@gmail.com \hspace{0.4cm} \{qind, jberent\}@google.com}}


\maketitle
\ificcvfinal\thispagestyle{empty}\fi

\begin{abstract}
Learning to localize and name object instances is a fundamental problem in vision, but state-of-the-art approaches rely on expensive bounding box supervision. While weakly supervised detection (WSOD) methods relax the need for boxes to that of image-level annotations, even cheaper supervision is naturally available in the form of unstructured textual descriptions that users may freely provide when uploading image content. However, straightforward approaches to using such data for WSOD wastefully discard captions that do not exactly match object names. Instead, we show how to squeeze the most information out of these captions by training a text-only classifier that generalizes beyond dataset boundaries. 
Our discovery provides an opportunity for learning detection models from noisy but more abundant and freely-available caption data.
We also validate our model on three classic object detection benchmarks and achieve state-of-the-art WSOD performance.
Our code is available at \url{https://github.com/yekeren/Cap2Det}.
\end{abstract}

\section{Introduction}
\label{sec:intro}

Learning to localize and classify visual is a fundamental problem in computer vision. It has a wide range of applications, including robotics, autonomous vehicles, intelligent video surveillance, and augmented reality. Since the renaissance of deep neural networks, object detection has been revolutionized by a series of groundbreaking works, including 
Faster-RCNN~\cite{ren2015faster}, Mask-RCNN~\cite{He_2017_ICCV} and YOLO \cite{redmon2016you}. 
Modern detectors can run in real-time on mobile devices, and have become the driving force for future technologies.

\begin{figure}[t]
    \centering
    \includegraphics[width=1.0\linewidth]{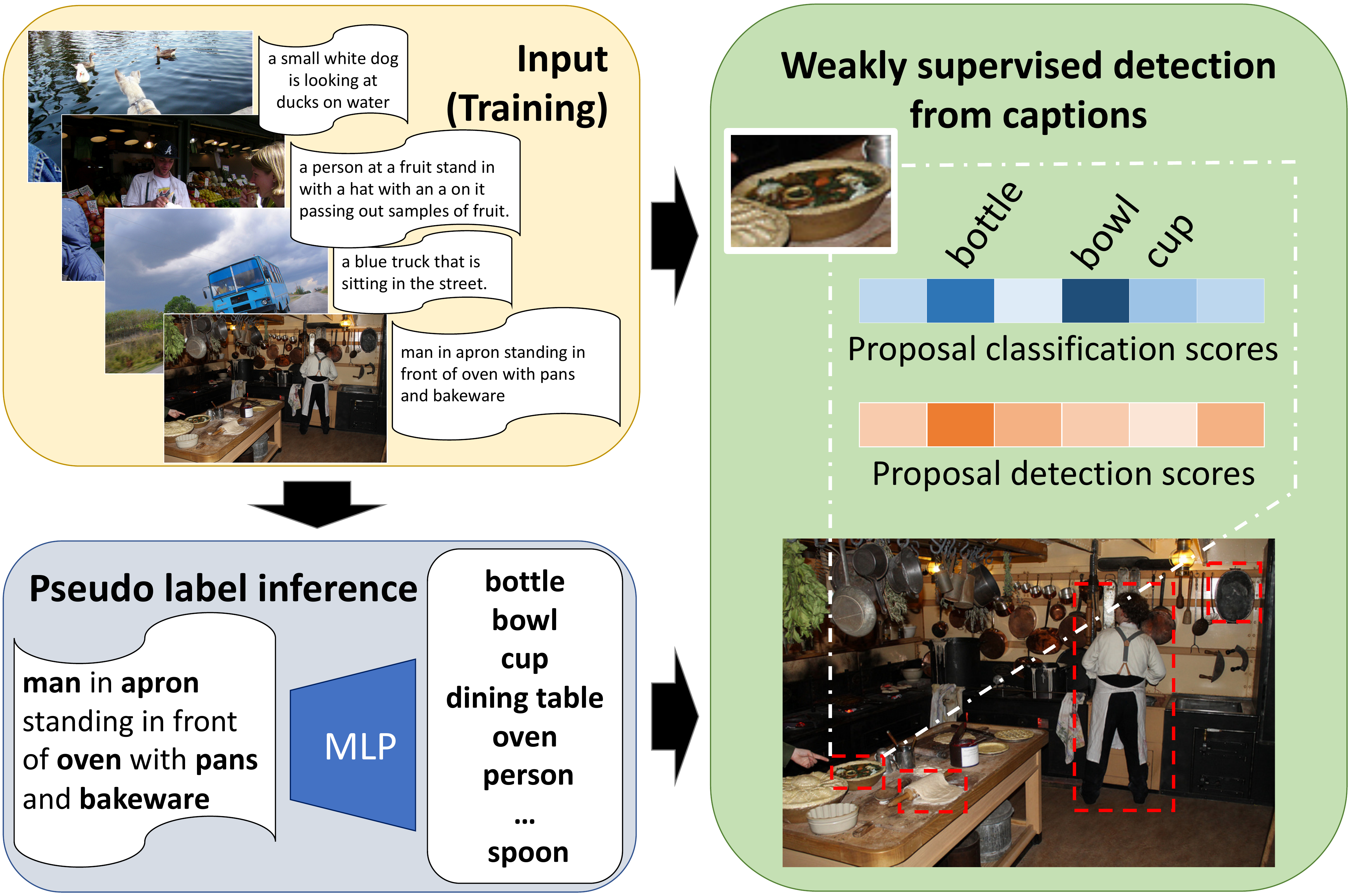}
    \caption{\textbf{An overview of our approach.} We leverage a novel source of weak supervision for object detection, namely natural language captions, by amplifying the signal these captions provide using a textual classifier. Even though ``bottle'', ``bowl'', ``cup'' and ``spoon'' are not mentioned in the caption, they are actually in the image, and our method correctly infers these image-level labels. We then train a model that scores object proposals in iterative fashion, to localize objects as its final output.}
    \label{fig:concept}
\vspace{-0.4cm}
\end{figure}

Despite these achievements, most modern detectors suffer from an important limitation: they are trained with expensive supervision in the form of large quantities of bounding boxes meticulously drawn by a large pool of human annotators. Due to the well-known domain shift problem \cite{chen2018domain,thomas2018artistic,li2016weakly,tang2012shifting,gopalan2011domain} and imperfect domain adaptation techniques, this means when detection is to be performed in a novel domain, the expensive annotation procedure needs to be repeated.
Weakly supervised object detection (WSOD) techniques aim to alleviate the burden of collecting such expensive box annotations.
The classic WSOD problem formulation~\cite{Bilen_2016_CVPR, Wei_2018_ECCV, Tang_2017_CVPR, tang2018pcl} treats an image as a bag of proposals, and learns to assign instance-level semantics to these proposals using multiple instance learning (MIL). WSOD has shown great potential for object detection, and the state-of-the-art model has reached 40\% mAP~\cite{tang2018pcl} on Pascal VOC 2012. However, one critical assumption in WSOD is that the image-level label should be \textit{precise}, indicating at least one proposal instance in the image needs to associate with the label. This assumption does not always hold especially for real-world problems and real-world supervision. 

Weakly supervised detection methods need large-scale image-level object category labels. These labels require human effort that is provided in an unnatural, crowdsourced environment. However, more natural supervision for objects exists---for example, in the form of natural language descriptions that web users provide when uploading their photos to social media sites such as YouTube or Instagram. 
There are tens of millions of photos uploaded to Instagram every day, and a majority of them have titles, tags, or descriptions. Abundant videos with subtitles are similarly available on YouTube.
These annotations are ``free'' in that no user was \emph{paid} to provide them; they arise out of innate needs of users to make their content available to others.

However, existing WSOD methods cannot use such supervision. First, these natural language descriptions are unstructured; they need to be parsed and words relevant for object recognition need to be extracted, while non-object words are removed. Second, these descriptions are both imprecise and non-exhaustive---they might mention content that is not in the image (e.g. what event the user was attending or who they met after the photo was taken), and also omit content that actually is in the image but is not interesting. Consider the image in the bottom-right of Fig.~\ref{fig:concept}. It contains numerous objects, many of which fairly large---e.g. dining table and bowls---yet the human providing the description did not mention these objects.
Thus, directly feeding web data to the state-of-the-art WSOD system contains numerous limitations---at the very least, it under-utilizes the rich supervision that captions can provide.

\begin{figure*}[t]
    \centering
    \includegraphics[width=1.0\linewidth]{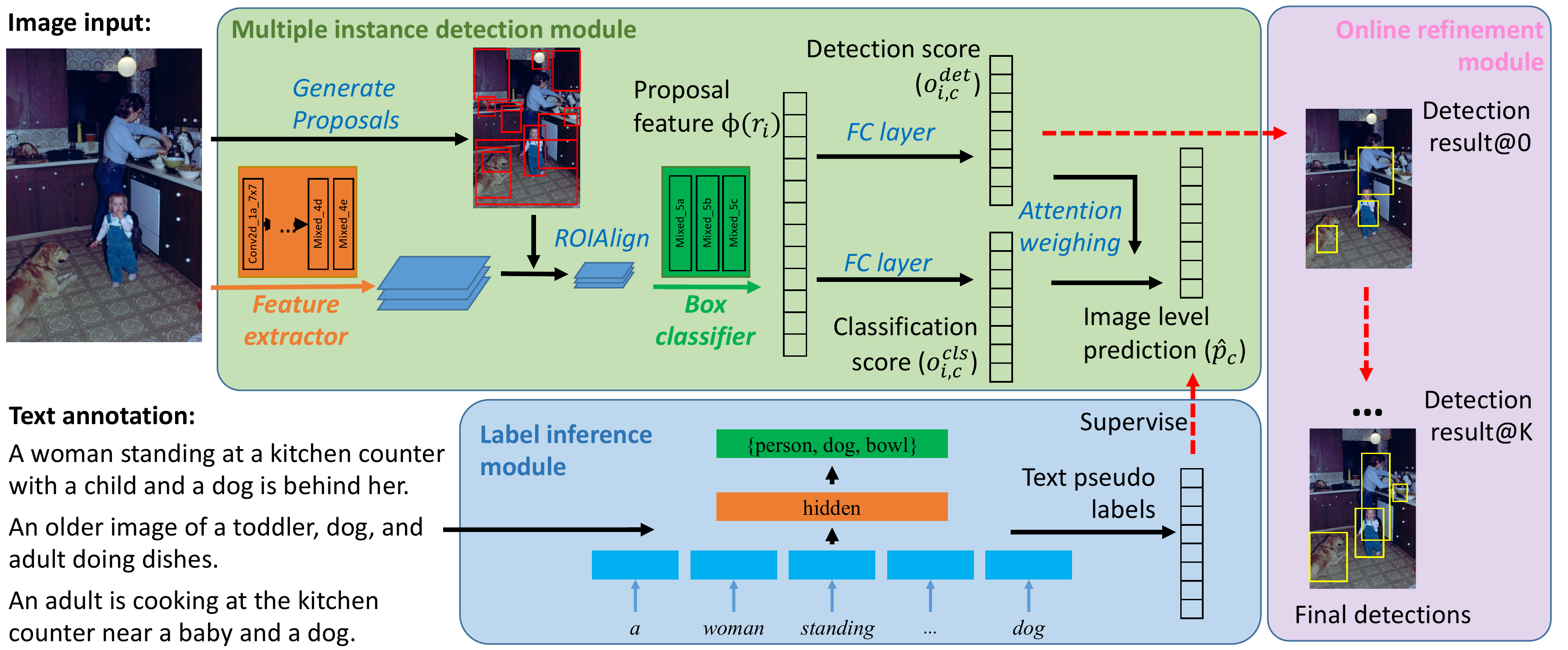}
    \caption{\textbf{Cap2Det: harvesting detection models from free-form text}. We propose to use a label inference module (bottom-center) to amplify signals in the free-formed texts to supervise the learning of the multiple instance detection network (top). The learned detection model is then refined by an online refinement module (right) to produce the final detection results.}
    \label{fig:architecture}
\vspace{-0.3cm}
\end{figure*}

To address this issue, we build an object detector from images paired with accompanying captions (sentences). 
Our model bridges human-written free-form texts and visual objects, and generates accurate bounding boxes over objects in an image. 
Our method relies on two key components.
First, we train a textual classifier to map captions to discrete object labels. This classifier is not dataset-dependent, requires only a small set of labels, and generalizes beyond dataset boundaries. It enables us to bridge the gap between what humans mention in a caption, and what truly is in an image. 
Second, we use the pseudo ground truth labels predicted by this textual classifier, to train a weakly supervised object detection method. The method we propose extracts region proposals off-the-shelf, then for each proposal and each class, learns both a class score and a detection score. These scores are then refined using an iterative approach, to produce final detection results.

The contributions of our paper are three-fold. 
First, we propose a new task of learning from noisy caption annotations, and set up a new benchmark. 
Rather than treating object categories as IDs only, we also leverage their semantics, as well as synonyms of those object names. 
Second, on this benchmark, we show that we outperform alternative uses of captions, e.g. exactly matching the captions to object category words, or retrieving hand-annotated or predicted synonyms to the object categories from the captions.
We show competitive WSOD performance by learning on COCO or Flickr30K captions.
We further validate the benefit of our COCO-trained text classifier by applying it on Flickr30K, and leveraging training on Flickr30K then evaluating on PASCAL. 
Finally, in a side-by-side comparison under the classic WSOD setting, our model demonstrates superior performance with image-level supervision and achieves state-of-the-art performance on all three WSOD benchmarks (48.5\% mAP@0.5 for VOC 2007, 45.1\% mAP@0.5 on VOC 2012 and 23.4\% mAP@0.5 on COCO).

The remainder of the paper is organized as follows. We overview related work in Sec.~\ref{sec:related}. 
In Sec.~\ref{sec:text_inference}, we discuss different ways to reduce the gap between free-form captions and object categories. 
In Sec.~\ref{sec:multiple_instance_detection}, we describe the backbone of our WSOD model, which combines prior work \cite{Tang_2017_CVPR,Szegedy_2016_CVPR} in a new way.
In Sec.~\ref{sec:results}, we compare our method to state-of-the-art and alternative methods. 
We conclude in Sec.~\ref{sec:conclusion}.
\section{Related Work}
\label{sec:related}

We formulate a new variant of weakly supervised object detection, where the supervision is even more weak than in prior work. We leverage captions, so we also discuss work that finds alignment between image regions and text.

\vspace{-0.6cm}
\paragraph{Weakly supervised detection via MIL.}

Most WSOD methods formulate the task as a multiple instance learning (MIL) problem. In this problem, proposals of an image are treated as a bag of candidate instances. If the image is labeled as containing an object, at least one of the proposals will be responsible to provide the prediction of that object. 
\cite{Oquab_2015_CVPR, Zhou_2016_CVPR} propose a Global Average (Max) Pooling layer to learn class activation maps. \cite{Bilen_2016_CVPR} propose  Weakly Supervised Deep Detection Networks (WSDDN) containing classification and detection data streams, where the detection stream weighs the results of the classification predictions. \cite{kantorov2016contextlocnet} improve WSDDN by considering context. \cite{Tang_2017_CVPR, tang2018pcl} jointly train multiple refining models together with WSDDN, and show the final model benefits from the online iterative refinement. \cite{Diba_2017_CVPR,Wei_2018_ECCV} apply a segmentation map and  \cite{Wei_2018_ECCV} incorporate saliency. Finally, \cite{Wan_2018_CVPR} adds a min-entropy loss to reduce the randomness of the detection results. Our work is similar to these since we also represent the proposals using a MIL weighted representation, however, we go one step further to successfully adopt a more advanced neural architecture, and a more challenging supervision scenario.

\paragraph{Learning from text.}

Recently there has been great interest in modeling the relationship between images and text, but to our knowledge, no work has explored learning a detector for images from captions. 
\cite{Chen_2017_CVPR} learn to discover and localize new objects from documentary \emph{videos} by associating subtitles to video tracklets. They extract keywords from the subtitles using TFIDF. Video provides benefits we cannot leverage, e.g. numerous frames containing nearly identical object instances. Importantly, we show that only using words that actually appear in the caption (as done with TFIDF) results in suboptimal performance compared to our method.
There is also work to associate phrases in the caption to visually depicted objects \cite{rohrbach2016grounding,hu2016segmentation,Anderson_2018_CVPR,Teney_2018_CVPR,Krause_2017_CVPR,Ye_2018_ECCV} but none enable training of an independent object detector with accurate localization and classification, as we propose. 

\vspace{-0.5cm}
\paragraph{Captions, categories and human bias.}

Our results show there is a gap between what humans name in captions, and what categorical annotations they provide. 
\cite{Misra_2016_CVPR} study a similar phenomenon they refer to as ``human reporting bias''.
They model the presence of an actual object as a latent variable, but we do the opposite---we model ``what's in the image'' by observing ``what's worth saying''.
Further, we use the resultant model as precise supervision to guide detection model learning.
In other work, \cite{Zhang_2018_BMVC} predict the nuance between an ad image and a slogan, \cite{Turakhia_2013_ICCV} study attribute dominance, and \cite{berg2012understanding} explore perceived visual importance.
\section{Approach}
\label{sec:approach}

The overall architecture of training using image-level text annotations is shown in Fig.~\ref{fig:architecture}. There are three major components: the label inference module, extracting the targets mentioned in noisy texts (bottom-center); the multiple instance detection module which estimates classification and detection scores for each proposal (top); and the online refinement module which refines these to get final detection results (right). They key novelty of our method lies in the label inference module.

\subsection{Label inference from text}
\label{sec:text_inference}

The foundation of WSOD builds on an important assumption from MIL (Eq.~\ref{eq:loss_mid}), which suggests that \textit{precise} image-level labels should be provided. However, gathering such \textit{clean} annotations is not trivial. In most real-life cases, the semantic counterpart for visual content appears in the form of natural language phrases, sentences, or even paragraphs (in newspapers), which is noisier than object labels. 

The straightforward solution of extracting object labels from captions via lexical matching, does not work well. Consider an image with three sentence descriptions:

``\textit {a \textbf{person} is riding a \textbf{bicycle} on the side of a bridge.}''

``\textit {a \textbf{man} is crossing the street with his \textbf{bike}.}''

``\textit {a \textbf{bicyclist} peddling down a busy city street.}''

However, only the first sentence exactly matches the categories ``person'' and ``bicycle''. Even if we allow synonyms of ``man'' and ``person'' or ``bicycle'' and ``bike'', only the first two precisely describe both objects, while the last one still misses the instance of ``bicycle'' unintentionally.

When using these examples to train object detectors, the first two instances may bring positive effect, but the last one will be wastefully discarded as false negative i.e. not relevant to the categories ``person'' or ``bicycle''.
Even worse, in the example shown in Fig.~\ref{fig:concept}, none of the captions (one shown) mention the ``bowls'' or ``spoons'' that are present, and only some mention the ``oven''.

This observation inspires us to amplify the supervision signal that captions provide, and squeeze more information out of them.
Fig.~\ref{fig:architecture} (bottom) shows the approach we use to amplify the signal. This text-only model takes free-form texts as input, embeds individual words to a 300D space using GloVe \cite{pennington2014glove}, and projects the embedded features to a 400D latent space. We then use max-pooling to aggregate the word-level representations. Then, we use this intermediate representation to predict the implied instances (e.g. 80 classes as defined in COCO, or any other categories); this prediction answers ``what's in the image'' and serves as pseudo image-level labels in training object detectors.

It is worth noting that there exists a subtle balance when using pseudo labels to train object detectors. Admittedly, our strategy increases the recall rates thus more data could be utilized. However, with the increased recall, precision will drop inevitably thus the fundamental assumption in MIL is threatened. Specifically, the \textit{precise label} assumption makes the model very sensitive to false positive cases: when inappropriate labels are given where none of the proposals have a good response, the model gets confused, resulting in non-optimal detections.

We finally adapt a two-steps procedure: first we look for an exact match of object labels from captions, following the intuition that \textit{explicitly mentioned objects} should be significant and obvious enough in the image; second, when no object can be matched, we use our label inference model to predict labels as \textit{unspoken intended objects} to guide the object detection. We show our method outperforms several strong alternatives that also infer pseudo labels.

\vspace{-0.5cm}
\paragraph{Discussion.} Our text classifier relies on both captions and category labels. However, once the bridge between captions and labels is established, this classifier generalizes to other datasets, as we show in Tab.~\ref{tab:result_cap_pascal_map}.
Importantly, we only need a small fraction of labels to train this text classifier; as we show in Fig.~\ref{fig:result_text_pr}, precision ranges between 89\% and 92\% when we use between only 5\% and 100\% of the COCO data, while recall is stable at 62\%.
Thus, our text model could learn from a \textit{single source} dataset with \textit{a few} labels, then it could transfer the knowledge to other \textit{target} datasets, requiring only free-form text as supervision.

\subsection{Detection from inferred labels}
\label{sec:multiple_instance_detection}

We next describe how we use the inferred pseudo labels to train an object detection model.
As shown in Fig.~\ref{fig:architecture}, we first extract proposals with accompanying features.
An image is fed into the pretrained (on ImageNet~\cite{imagenet_cvpr09}) convolutional layers. 
Then, \textit{ROIAlign}~\cite{He_2017_ICCV} is used for cropping the proposals (at most 500 boxes per image) generated by \textit{Selective Search}~\cite{uijlings2013selective}, resulting in fixed-sized convolutional feature maps.
Finally, a box feature extractor 
is applied to extract a fixed-length feature for each proposal.
If $[r_1,\dots,r_m]$ are the proposals 
of a given image $x$, 
this process results in proposal feature vectors $[\phi(r_1),\dots, \phi(r_m)]$ where each $\phi(r_i)\in \mathbb{R}^d$. 
Note that while our model is pretrained on ImageNet, it \emph{does not leverage} any image labels at all on the datasets on which we train and evaluate our detection models (PASCAL and COCO).

\vspace{-0.2cm}
\subsubsection{Weakly supervised detection}
We next introduce the prediction of image-level labels $\hat{p}_c$ ($c\in \{1,\dots,C\}$, where $C$ is the number of classes) and of detection scores as a by-product.
The proposal features $\phi(r_i)$ are fed into two parallel fully-connected layers to compute the detection scores $o_{i,c}^\text{det}\in \mathbb{R}^1$ (top branch in the green MIL module in Fig.~\ref{fig:architecture}) and classification scores $o_{i,c}^\text{cls}\in \mathbb{R}^1$ (bottom branch), in which both scores are related to a specific class $c$ and the particular proposal $r_i$: 
\vspace{-0.2cm}
\begin{equation} \label{eq:cls_score}
    o_{i,c}^\text{cls}=w_c^{\text{cls}\intercal}\phi(r_i)+b_c^\text{cls}, ~~~~~
    o_{i,c}^\text{det}=w_c^{\text{det}\intercal}\phi(r_i)+b_c^\text{det}
\vspace{-0.2cm}
\end{equation}

We convert these scores 
into: (1) $p_{i,c}^\text{cls}$, 
the probability that object $c$ presents in proposal $r_i$; and (2) $p_{i,c}^\text{det}$, 
the probability that $r_i$ is important for predicting image-level label $y_c$:
\vspace{-0.2cm}
\begin{equation} \label{eq:cls_proba}
    p_{i,c}^\text{cls}=\sigma(o_{i,c}^\text{cls}), ~~~~~
    p_{i,c}^\text{det}=\frac{\exp(o_{i,c}^\text{det})}{\sum_{j=1}^{m}\exp(o_{j,c}^\text{det})}
\vspace{-0.2cm}
\end{equation}

Finally, the aggregated image-level prediction is computed as follows, 
where greater values of $\hat{p}_c \in [0,1]$ mean higher likelihood that $c$ is present in the image:
\vspace{-0.2cm}
\begin{equation} \label{eq:final_proba}
    \hat{p}_c=\sigma\bigg(\sum\limits_{i=1}^{m} p_{i,c}^\text{det} o_{i,c}^\text{cls}\bigg)
\vspace{-0.2cm}
\end{equation}

Assuming the label $y_c=1$ if and only if class $c$ is present,
the \textbf{m}ultiple \textbf{i}nstance \textbf{d}etection loss used for training the model is defined as:
\vspace{-0.2cm}
\begin{equation} \label{eq:loss_mid}
    L_\text{mid}=- \sum_{c=1}^C \bigg [
        y_c\log \hat{p}_c + (1-y_c)\log(1-\hat{p}_c) 
    \bigg]
\vspace{-0.2cm}
\end{equation}

\vspace{-0.5cm}
\paragraph{Preliminary detection scores.}
The weakly supervised detection score given both proposal $r_i$ and class $c$ is the product of $p_{i,c}^\text{cls}$ and $p_{i,c}^\text{det}$ which is further refined as described in Sec.~\ref{sec:online_instance_classifier_refinement}.

\vspace{-0.2cm}
\subsubsection{Online instance classifier refinement}
\label{sec:online_instance_classifier_refinement}

The third component of our WSOD model is Online Instance Classifier Refinement (OICR), as proposed by Tang \etal~\cite{Tang_2017_CVPR}. 
The main idea behind OICR is simple: Given a ground-truth class label, the top-scoring proposal, as well as proposals highly overlapping with it, are selected as references. These proposals are treated as positive examples for training the box classifier of this class while others are treated as negatives.
The initial top-scoring proposal may only partially cover the object, so allowing highly-overlapped proposals to be treated as positives gives them a second chance to be considered as containing an object, in the subsequent model refinement. 
This reduces the chance of propagating incorrect predictions. In addition, sharing the convolutional features between the original and refining models makes the training more robust.

Following \cite{Tang_2017_CVPR}, we stack multiple refining classifiers and use the output of the previous one to generate instance-level supervision to train the successor. 
The detection score at the $0$-th iteration is computed using  $s_{i,c}^{(0)}=p_{i,c}^{cls}p_{i,c}^{det}, s_{i,C+1}^{(0)}=0$ (where $C+1$ is the background class).
Given the detection score $s_{i,c}^{(k)}$ at the $k$-th iteration, we use the image-level label to get the \textit{instance-level} supervision $y_{i,c}^{(k+1)}$ at the $(k+1)$-th iteration. Assume that $c'$ is a label attached to image $x$, we first look for the top-scoring box $r_j$ ($j=\text{arg}\max\limits_i s_{i,c'}^{(k)}$).
We then 
let $y_{i,c'}^{(k+1)}=1, \forall i \in \{l|\textit{IoU}(r_l, r_j)>\textit{threshold}\}$. 
When $k>0$, $s_{i,c}^{(k)}$ is inferred using a $(C+1)$-way FC layer, as in Eq.~\ref{eq:cls_score}.
The OICR training loss is defined in Eq.~\ref{eq:loss_oicr}.
\vspace{-0.2cm}
\begin{equation} \label{eq:loss_oicr}
    L_\text{oicr}^k=- \frac{1}{m} \sum_{i=1}^{m}\sum_{c=1}^{C+1} 
        \hat{y}_{i,c}^{(k)} \log s_{i,c}^{(k)}, ~~~~ k=1,\dots,K
\vspace{-0.2cm}
\end{equation}

Unlike the original OICR, our WSOD module aggregates logits instead of probability scores, which in our experience stabilizes training. We also removed the reweighing of untrustworthy signals emphasized in \cite{Tang_2017_CVPR} since we found it did not contribute significantly.

The final loss we optimize is Eq.~\ref{eq:loss}. We refine our model for 3 times ($K=3$) if not mentioned otherwise.
\vspace{-0.2cm}
\begin{equation} \label{eq:loss}
    L=L_\text{mid}+\sum_{k=1}^{K}L_\text{oicr}^k
\vspace{-0.2cm}
\end{equation}
\vspace{-0.5cm}
\section{Experiments}
\label{sec:results}

We evaluate all components of our method: the text classifier that learns to map captions to object labels, the weakly supervised detection module, and the refinement. We show that compared to alternative strategies, our approach extracts the most accurate and expansive information from the captions (Sec.~\ref{sec:results_using_caption}). By training on COCO captions, we achieve close to state-of-the-art results on weakly supervised detection on PASCAL, even though the supervision we leverage is weaker than competitor methods. Importantly, our text classifier allows us to excel at the task of training on Flickr30K to detect on PASCAL, even though that classifier was trained on a different dataset (COCO). Finally, we show our approach outperforms prior methods on the task of learning from image-level labels (Sec.~\ref{sec:results_using_label}).

\subsection{Implementation details}
\label{sec:details}

Before training the detector, we use Selective Search \cite{uijlings2013selective} from OpenCV \cite{opencv_library} to extract at most 500 proposals for each image. We follow the ``Selective search quality'' parameter settings in \cite{uijlings2013selective}. 
We prefer Selective Search because it is a generic, dataset-independent proposal generation procedure, as opposed to other CNN-based alternatives which are trained end-to-end from a specific dataset in a supervised fashion. We also experimented with Edge Boxes \cite{zitnick2014edge} but got inferior performance.
We use TensorFlow \cite{Abadi:2016:TSL:3026877.3026899} as our training framework. To compute the proposal feature vectors, we use the layers (``Conv2d\_1a\_7x7'' to ``Mixed\_4e'') from Inception-V2 \cite{Szegedy_2016_CVPR} to get the conv feature map, and the layers (``Mixed\_5a'' to ``Mixed\_5c'') from the same model to extract the proposal feature vectors after the ROIAlign \cite{He_2017_ICCV} operation.
The Inception-V2 model is pretrained on  ImageNet \cite{imagenet_cvpr09}; the supervised detector counterpart of our model, using this architecture, was explored by \cite{Huang_2017_CVPR}.
To augment the training data, we resize the image randomly to one of the four scales $\{400, 600, 800, 1200\}$.
We also randomly flip the image left to right at training time. At test time, we average the proposal scores from the different resolution inputs.
We set the number of refinements to $3$ for the OICR since it gives the best performance. For post-processing, we use non-maximum-suppression with IoU threshold of $0.4$. We use the AdaGrad optimizer, a learning rate of $0.01$, and a batch size of 2 as commonly used in WSOD methods~\cite{Tang_2017_CVPR, tang2018pcl}. 
The models are usually trained for 100K iterations on Pascal VOC (roughly 40 epochs on VOC2007 and 17 epochs on VOC2012) and 500K on COCO (8.5 epochs), using a validation set to pick the best model.
Our implementation is available at \url{https://github.com/yekeren/Cap2Det}.

\subsection{Using captions as supervision}
\label{sec:results_using_caption}

\begin{table*}[t]
    \footnotesize
    \centering
    \setlength\tabcolsep{0pt} 
    \begin{tabularx}{\textwidth}{p{3.0cm}|*{20}{>{\centering\arraybackslash}X}|>{\centering\arraybackslash}X}
    \toprule
        Methods & \rotatebox{90}{aero} & \rotatebox{90}{bike} & \rotatebox{90}{bird} & \rotatebox{90}{boat} & \rotatebox{90}{bottle} & \rotatebox{90}{bus} & \rotatebox{90}{car} & \rotatebox{90}{cat} & \rotatebox{90}{chair} & \rotatebox{90}{cow} & \rotatebox{90}{table} & \rotatebox{90}{dog} & \rotatebox{90}{horse} & \rotatebox{90}{mbike} & \rotatebox{90}{person} & \rotatebox{90}{plant} & \rotatebox{90}{sheep} & \rotatebox{90}{sofa} & \rotatebox{90}{train} & \rotatebox{90}{tv} & \rotatebox{90}{mean} \\
    \Xhline{2\arrayrulewidth}
        \multicolumn{21}{l}{Training on different datasets using ground-truth labels:} \rule{0pt}{1.01em} & \\
        \textsc{GT-Label VOC} & 68.7 & 49.7 & 53.3 & 27.6 & 14.1 & 64.3 & 58.1 & 76.0 & 23.6 & 59.8 & 50.7 & 57.4 & 48.1 & 63.0 & 15.5 & 18.4 & 49.7 & 55.0 & 48.4 & 67.8 & 48.5 \\ 
        \textsc{GT-Label COCO} & 65.3 & 50.3 & 53.2 & 25.3 & 16.2 & 68.0 & 54.8 & 65.5 & 20.7 & 62.5 & 51.6 & 45.6 & 48.6 & 62.3 & 7.2 & 24.6 & 49.6 & 34.6 & 51.1 & 69.3 & 46.3 \\
    \hline
        \multicolumn{21}{l}{Training on COCO dataset using captions:} \rule{0pt}{1.01em} & \\
        \textsc{ExactMatch (EM)} & 63.0 & \textbf{50.3} & 50.7 & 25.9 & \textbf{14.1} & 64.5 & 50.8 & 33.4 & 17.2 & 49.0 & 48.2 & 46.7 & 44.2 & 59.2 & 10.4 & 14.3 & 49.8 & 37.7 & 21.5 & 47.6 & 39.9 \\ 
        \textsc{EM + GloVePseudo} & \textbf{66.6} & 43.7 & 53.3 & 29.4 & 13.6 & 65.3 & \textbf{51.6} & 33.7 & 15.6 & 50.7 & 46.6 & 45.4 & 47.6 & \textbf{62.1} & 8.0 & \textbf{15.7} & 48.6 & 46.3 & 30.6 & 36.4 & 40.5 \\
        \textsc{EM + LearnedGloVe} & 64.1 & 49.9 & \textbf{58.6} & 24.9 & 13.2 & \textbf{66.9} & 49.2 & 26.9 & 13.1 & 57.7 & \textbf{52.8} & 42.6 & \textbf{53.2} & 58.6 & 14.3 & 15.0 & 45.2 & 50.3 & 34.1 & 43.5 & 41.7 \\
        \textsc{EM + ExtendVocab} & 65.0 & 44.9 & 49.2 & \textbf{30.6} & 13.6 & 64.1 & 50.8 & 28.0 & \textbf{17.8} & 59.8 & 45.5 & \textbf{56.1} & 49.4 & 59.1 & 16.8 & 15.2 & \textbf{51.1} & \textbf{57.8} & 14.0 & \textbf{61.8} & 42.5 \\ 
        \textsc{EM + TextClsf} & 63.8 & 42.6 & 50.4 & 29.9 & 12.1 & 61.2 & 46.1 & \textbf{41.6} & 16.6 & \textbf{61.2} & 48.3 & 55.1 & 51.5 & 59.7 & \textbf{16.9} & 15.2 & 50.5 & 53.2 & \textbf{38.2} & 48.2 & \textbf{43.1} \\
    \hline
        \multicolumn{21}{l}{Training on Flickr30K dataset using captions:} \rule{0pt}{1.01em} & \\
        \textsc{ExactMatch (EM)} & \textbf{46.6} & \textbf{42.9} & 42.0 & 9.6 & 7.7 & 31.6 & 44.8 & 53.2 & 13.1 & 28.0 & 39.1 & 43.2 & 31.9 & \textbf{52.5} & 4.0 & \textbf{5.1} & 38.0 & 28.7 & \textbf{15.8} & 41.1 & 31.0 \\
        \textsc{EM + ExtendVocab} & 37.8 & 37.6 & 35.5 & 11.0 & \textbf{10.3} & 18.0 & 47.9 & 51.3 & \textbf{17.7} & 25.5 & 37.0 & 47.9 & \textbf{35.2} & 46.1 & \textbf{15.2} & 0.8 & 27.8 & 35.6 & 5.8 & 42.0 & 29.3 \\
        \textsc{EM + TextClsf} & 24.1 & 38.8 & \textbf{44.5} & \textbf{13.3} & 6.2 & \textbf{38.9} & \textbf{49.9} & \textbf{60.4} & 12.4 & \textbf{47.4} & \textbf{39.2} & \textbf{59.3} & 34.8 & 48.1 & 10.7 & 0.3 & \textbf{42.4} & \textbf{39.4} & 14.1 & \textbf{47.3} & \textbf{33.6} \\
    \bottomrule
    \end{tabularx}
    \vspace{-0.2cm}
    \caption{\textbf{Average precision (in \%) on the VOC 2007 test set (learning from COCO and Flickr30K captions)}. We learn the detection model from the COCO captions describing the 80 objects, but evaluate on only the overlapping 20 VOC objects.}
    \label{tab:result_cap_pascal_map}
\end{table*}

In this section, we evaluate our full method, including our proposal for how to squeeze the most information out of the weak supervision that captions provide (Sec.~\ref{sec:text_inference}).
We also experiment with alternative strategies of generating pseudo labels, and evaluate the performance in terms of precision and recall by comparing with ground-truth labels.

\vspace{-0.5cm}
\paragraph{Alternative strategies.} 
We compared with multiple pseudo-label generation baselines when lexical matching (\textsc{ExactMatch}) fails to find a match. As previous examples show, considering synonyms 
can effectively reduce off-target matching rates. Thus our first baseline adopts a \emph{manually constructed, hence expensive} COCO synonym vocabulary list (\textsc{ExtendVocab}) which maps 413 words to 80 categories \cite{Lu_2018_CVPR}. Another variant, \textsc{GloVePseudo}, takes advantage of GloVe word embeddings \cite{pennington2014glove}, assigning pseudo-labels for a sentence by looking for the category that has the smallest embedding distance to any word in the sentence. We also follow a similar strategy with ~\cite{Ye_2018_ECCV} to finetune the GloVe word embeddings on COCO using a visual-text ranking loss, and use the pseudo labels retrieved by the resultant \textsc{LearnedGloVe} as a stronger baseline.
The final reference model of using ground-truth image-level labels \textsc{GT-Label} is an upper bound.
Note that apart from the strategy used to mine image-level labels, these strategies all use the same architecture and WSOD approach as our method (Sec.~\ref{sec:multiple_instance_detection}).
In later sections, we show combinations of the exact match strategy with these methods (when exact match fails), resulting in \textsc{EM+GloVePseudo}, \textsc{EM+LearnedGloVe}, \textsc{EM+ExtendVocab} and \textsc{EM+TextClsf}.
We examine how well these and other strategies leverage captions from COCO and Flickr30K~\cite{young2014image} to produce accurate detection. 

\begin{figure}[t]
    \centering
    \includegraphics[width=1.0\linewidth]{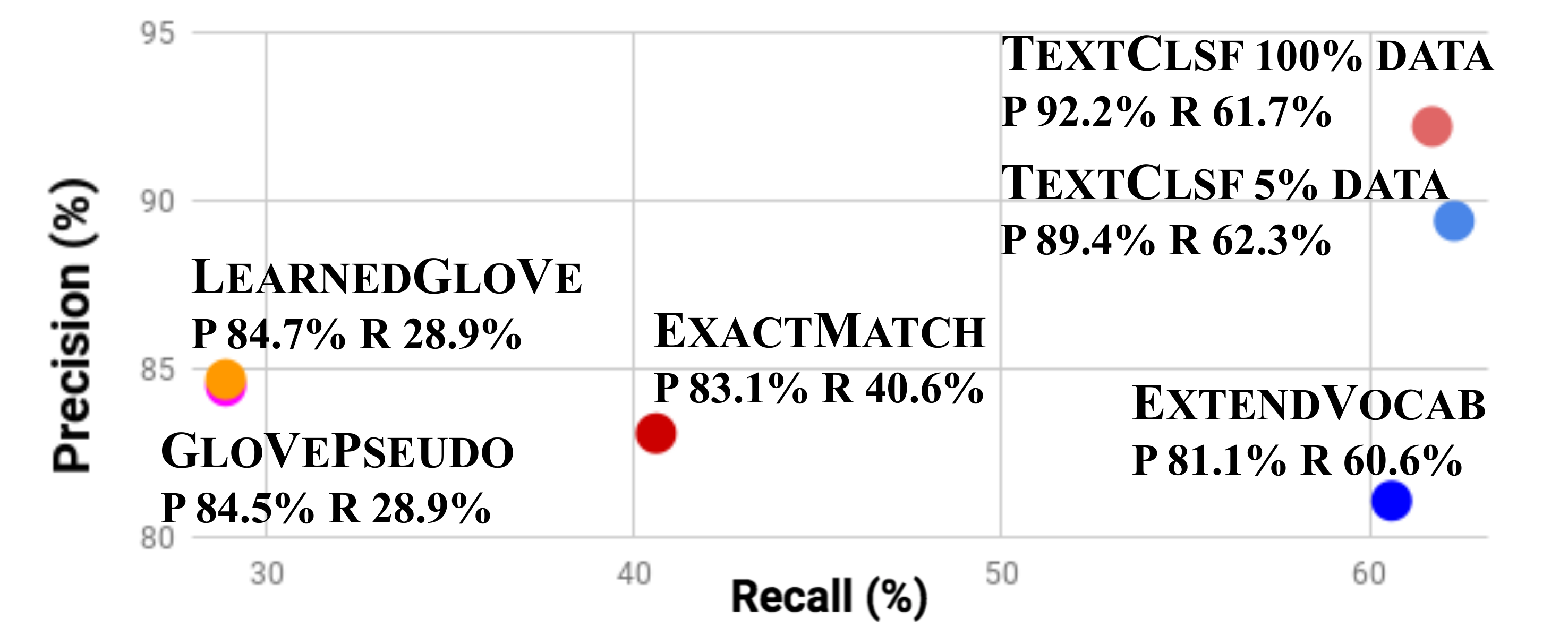}
\vspace{-0.7cm}
    \caption{\textbf{Analysis of different text supervision}. We compare the pseudo labels (Sec.~\ref{sec:text_inference}) to COCO \textit{val} ground-truth.}
    \label{fig:result_text_pr}
\vspace{-0.5cm}
\end{figure}

\vspace{-0.5cm}
\paragraph{Analysis of textual supervision.}

In Fig.~\ref{fig:result_text_pr} we show the \textit{precision} and \textit{recall} of these label inference methods evaluated directly on the COCO image-level labels (5,000 examples of the \textit{val2017} set). We observe that \textsc{ExtendVocab}, which uses the hand-crafted word-synonyms dictionary, provides the best recall (60.6\%) among all methods but provides the worst precision of 81.1\%. The word-embedding-based top-scoring matching methods of \textsc{GloVePseudo} and \textsc{LearnedGloVe} provide precise predictions (84.5\% and 84.7\% respectively, which are the highest). 
However, our \textsc{TextClsf} achieves significantly improved precision compared to these. 
We would like to point out that while in Tab.~\ref{tab:result_cap_pascal_map} and \ref{tab:result_cap_coco_map}, our method uses the full COCO training set (118,287 concatenated captions), it achieves very similar performance with even a small fraction of the data. With 5\% of the data, the method achieves 89\% precision (vs 92\% precision with 100\% of the data), both of which are much higher than any other baselines; recall is about 62\% for both 5\% and 100\% training data.
In other words, it is sufficient to use a small portion of precise text labels to train a  generalizable label inference classifier, and the knowledge can transfer to other datasets as we show in Tab.~\ref{tab:result_cap_pascal_map}.

\begin{figure}[t]
\vspace{-0.4cm}
    \centering
    \includegraphics[width=1.0\linewidth]{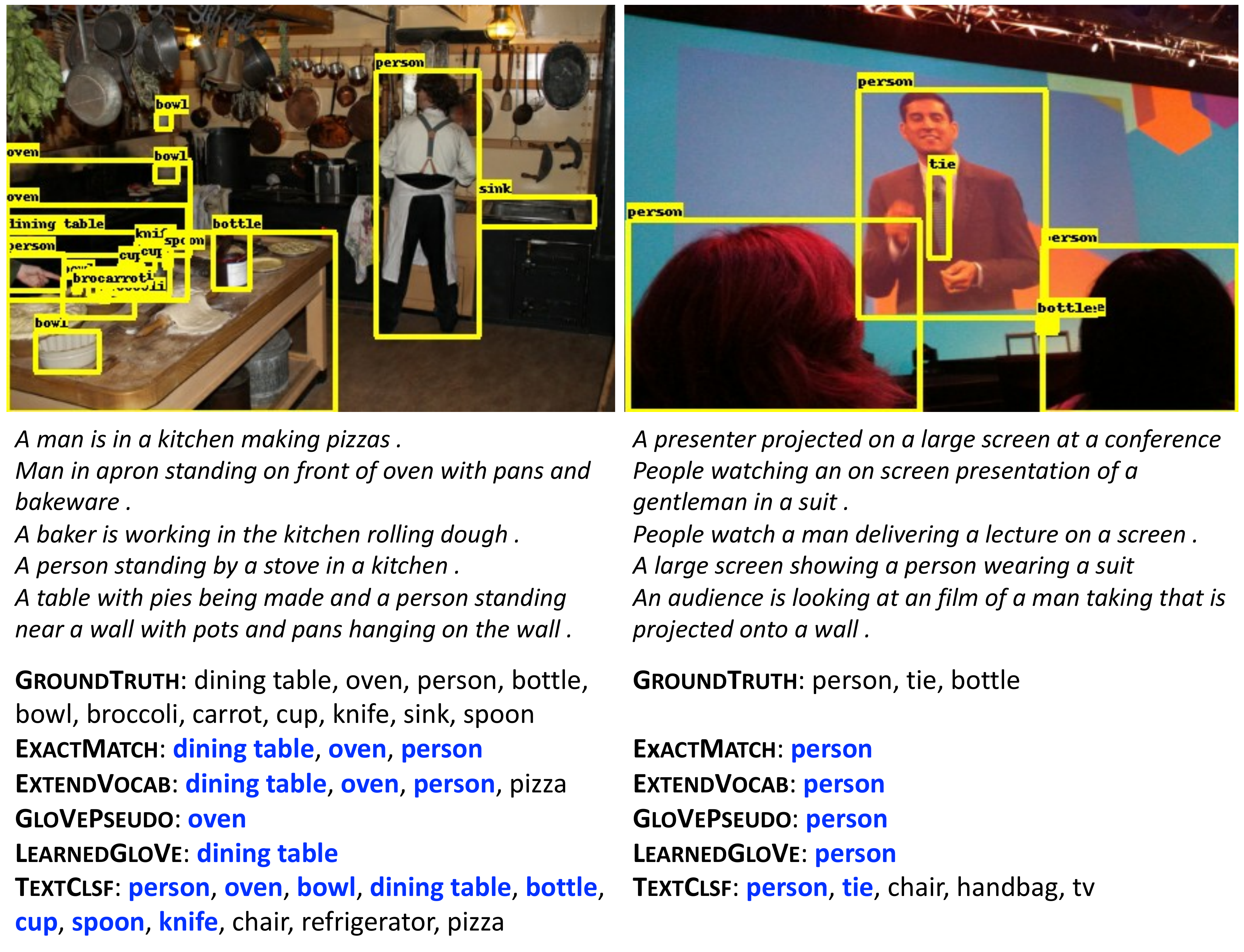}
\vspace{-0.6cm}
    \caption{\textbf{Demonstration of different pseudo labels}. Our method fills the gap between what is present and what is mentioned, by making inferences on the semantic level. Matches to the ground truth are shown in blue.}
    \label{fig:result_pseudo_labels}
\end{figure}

To better understand the generated labels, we show two qualitative examples in Fig.~\ref{fig:result_pseudo_labels}. The image on the right shows that our model infers ``tie'' from the observation of ``presenter'', ``conference'' and ``suit'', while all other methods fail to extract this object category for visual detection. 
We argue the capability of inferring reasonable labels from captions is critical for learning detection model from noisy captions.

\vspace{-0.5cm}
\paragraph{Training with COCO captions.}
We next train our detection model using the COCO captions~\cite{chen2015microsoft}. We use the 591,435 annotated captions paired to the 118,287 \textit{train2017} images.
For evaluation, we use the COCO \textit{test-dev}2017 and PASCAL VOC 2007 \textit{test} sets.
In our supplementary file, we show qualitative examples from the COCO \textit{val} set.

Tab.~\ref{tab:result_cap_pascal_map} shows the results on PASCAL VOC 2007. At the top are two upper-bound methods that train on image-level \emph{labels}, while the rest of the methods train on image-level \emph{captions}.
\textsc{ExactMatch} (\textsc{EM}) performs the worst probably due to its low data utilization rate, as evidenced by the fact that all methods incorporating pseudo labels improve performance notably.
Specifically, \textsc{EM+GlovePseudo} uses free knowledge of the pre-trained GloVe embeddings. It alleviates the synonyms problem to a certain extent, thus it improves the mAP by 2\% compared to \textsc{ExactMatch}. However, the GloVe embedding is not optimized for the specific visual-captions, resulting in noisy knowledge transformation.
\textsc{EM+LearnedGloVe} learns dataset-specific word embeddings. Its performance, as expected, is 3\% better than \textsc{EM+GloVePseudo} in terms of mAP.
The strongest baseline is \textsc{EM+ExtendVocab}, as the manually picked vocabulary list covers most frequent occurrences. However, collecting such vocabulary requires human effort, and is not a scalable and transferable strategy. 
Our \textsc{EM+TextClsf} outperforms this expensive baseline, especially for categories ``cat'', ``cow'', ``horse'', and ``train''.

\begin{table}[t]
    \footnotesize
    \centering
    \setlength\tabcolsep{0pt} 
    \begin{tabularx}{\linewidth}{>{\hsize=2.5\hsize}X|*{3}{>{\hsize=0.75\hsize\centering\arraybackslash}X}|*{3}{>{\hsize=0.75\hsize\centering\arraybackslash}X}}
    \toprule
        \multirow{2}{*}{Methods} & \multicolumn{3}{c|}{Avg. Precision, IoU} & \multicolumn{3}{c}{Avg. Precision, Area} \\
        & 0.5:0.95 & 0.5 & 0.75 & S & M & L \\
    \midrule
        \textsc{GT-Label} & 10.6 & 23.4 & 8.7 & 3.2 & 12.1 & 18.1 \\
    \hline
        \textsc{ExactMatch (EM)} \rule{0px}{1.01em} & 8.9 & 19.7 & 7.1 & 2.3 & 10.1 & 16.3 \\
        \textsc{EM + GloVePseudo} & 8.6 & 19.0 & 6.9 & 2.2 & 10.0 & 16.0 \\
        \textsc{EM + LearnedGloVe} & 8.9 & 19.7 & 7.2 & 2.5 & 10.4 & \textbf{16.6} \\
        \textsc{EM + ExtendVocab} & 8.8 & 19.4 & 7.1 & 2.3 & 10.5 & 16.1 \\
        \textsc{EM + TextClsf}& \textbf{9.1} & \textbf{20.2} & \textbf{7.3} & \textbf{2.6} & \textbf{10.8} & \textbf{16.6} \\
    \bottomrule
    \end{tabularx}
    \caption{\textbf{COCO test-dev results (learning from COCO captions)}. We report these numbers by submitting to the COCO evaluation server. The best method is shown in \textbf{bold}.}
    \label{tab:result_cap_coco_map}
\vspace{-0.5cm}
\end{table}

At the top of Tab.~\ref{tab:result_cap_pascal_map} are two upper-bound methods which rely on ground-truth image-level captions. Despite the noisy supervision, our \textsc{EM+TextClsf} almost bridges the gap to the COCO-labels upper bound.

For the results on COCO (Tab.~\ref{tab:result_cap_coco_map}), the gaps in performance between the different methods are smaller, but as before, our proposed \textsc{EM+TextClsf} shows the best performance. We believe the smaller gaps are because many of the COCO objects are not described precisely via natural language, and the dataset itself is more challenging than PASCAL thus gain may be diluted by tough examples.

\vspace{-0.5cm}
\paragraph{Training with Flickr30K captions.}
We also train our model on the Flickr30K~\cite{young2014image} dataset, which contains 31,783 images and 158,915 descriptive captions.
Training on Flickr30K is more challenging: on one hand, it includes less data compared to COCO; on the other hand, we observe that the recall rate of the captions is only 48.9\% with \textsc{ExactMatch} 
which means only half of the data can be matched to some class names.
The results are  shown in the bottom of Tab.~\ref{tab:result_cap_pascal_map}.
We observe that due to the limited training size, the detection models trained on Flickr30K captions achieve weaker performance than those trained on COCO captions. However, given the ``free'' supervision, the 33.6\% mAP is still very encouraging. Importantly, we observe that even though our text classifier is trained on COCO captions and labels, it generalizes well to Flickr30K captions, as evidenced by the gap between \textsc{EM+TextClsf} and \textsc{EM+ExtendVocab}. 

\begin{figure}[t]
\vspace{-0.5cm}
    \centering
    \includegraphics[width=1.0\linewidth]{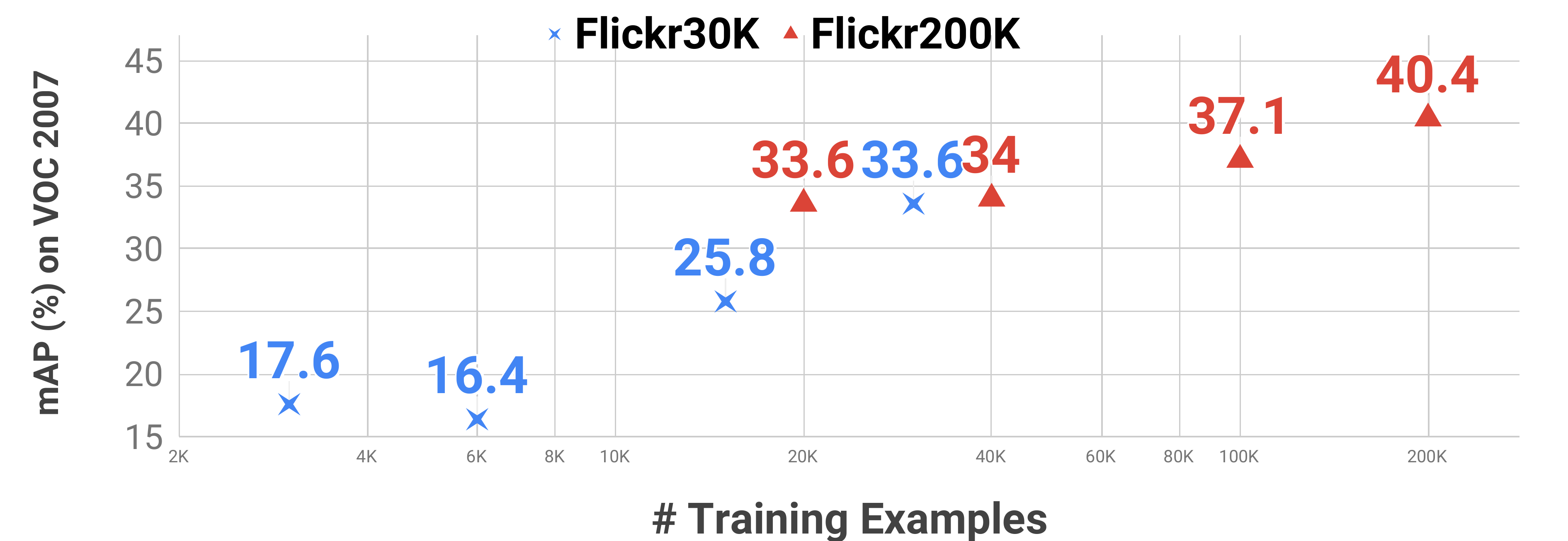}
    \vspace{-0.7cm}
    \caption{\textbf{Data vs. Performance}. Our text classifier learned on COCO generalized well on Flickr30K and the noisier Flickr200K data formed by user-generated content tags. }
    \label{fig:data_vs_map}
\vspace{-0.2cm}
\end{figure}

\begin{table*}[t]
    \footnotesize
    \centering
    \setlength\tabcolsep{0pt} 
    \begin{tabularx}{\textwidth}{p{3.6cm}|*{20}{>{\centering\arraybackslash}X}|>{\centering\arraybackslash}X}
    \toprule
        Methods & \rotatebox{90}{aero} & \rotatebox{90}{bike} & \rotatebox{90}{bird} & \rotatebox{90}{boat} & \rotatebox{90}{bottle} & \rotatebox{90}{bus} & \rotatebox{90}{car} & \rotatebox{90}{cat} & \rotatebox{90}{chair} & \rotatebox{90}{cow} & \rotatebox{90}{table} & \rotatebox{90}{dog} & \rotatebox{90}{horse} & \rotatebox{90}{mbike} & \rotatebox{90}{person} & \rotatebox{90}{plant} & \rotatebox{90}{sheep} & \rotatebox{90}{sofa} & \rotatebox{90}{train} & \rotatebox{90}{tv} & \rotatebox{90}{mean} \\
    \midrule
        \multicolumn{22}{l}{VOC 2007 results:} \\
        OICR VGG16~\cite{Tang_2017_CVPR} & 58.0 & 62.4 & 31.1 & 19.4 & 13.0 & \textbf{65.1} & 62.2 & 28.4 & 24.8 & 44.7 & 30.6 & 25.3 & 37.8 & 65.5 & 15.7 & \textbf{24.1} & 41.7 & 46.9 & \textbf{64.3} & 62.6 & 41.2 \\
        PCL-OB-G VGG16~\cite{tang2018pcl} & 54.4 & \textbf{69.0} & 39.3 & 19.2 & \textbf{15.7} & 62.9 & \textbf{64.4} & 30.0 & \textbf{25.1} & 52.5 & 44.4 & 19.6 & 39.3 & \textbf{67.7} & \textbf{17.8} & 22.9 & 46.6 & \textbf{57.5} & 58.6 & 63.0 & 43.5 \\
        TS$^2$C~\cite{Wei_2018_ECCV} & 59.3 & 57.5 & 43.7 & 27.3 & 13.5 & 63.9 & 61.7 & 59.9 & 24.1 & 46.9 & 36.7 & 45.6 & 39.9 & 62.6 & 10.3 & 23.6 & 41.7 & 52.4 & 58.7 & 56.6 & 44.3 \\
    \hline
        OICR Ens.+FRCNN~\cite{Tang_2017_CVPR} & 65.5 & 67.2 & 47.2 & 21.6 & 22.1 & 68.0 & 68.5 & 35.9 & 5.7 & \textit{63.1} & 49.5 & 30.3 & \textit{64.7} & 66.1 & 13.0 & \textit{25.6} & \textit{50.0} & 57.1 & 60.2 & 59.0 & 47.0 \\
        PCL-OB-G Ens.+FRCNN~\cite{tang2018pcl} & 63.2 & \textit{69.9} & 47.9 & 22.6 & \textit{27.3} & \textit{71.0} & \textit{69.1} & 49.6 & 12.0 & 60.1 & \textit{51.5} & 37.3 & 63.3 & 63.9 & 15.8 & 23.6 & 48.8 & 55.3 & 61.2 & 62.1 & \textit{48.8} \\
    \hline
        \textbf{Ours} \rule{0pt}{1.01em} & \textbf{68.7} & 49.7 & \textbf{53.3} & \textbf{27.6} & 14.1 & 64.3 & 58.1 & \textbf{76.0} & 23.6 & \textbf{59.8} & \textbf{50.7} & \textbf{57.4} & \textbf{48.1} & 63.0 & 15.5 & 18.4 & \textbf{49.7} & 55.0 & 48.4 & \textbf{67.8} & \textbf{48.5} \\ 
    \Xhline{2\arrayrulewidth}
        \multicolumn{21}{l}{VOC 2012 results:} & \rule{0pt}{1.1em} \\
        OICR VGG16~\cite{Tang_2017_CVPR} & 67.7 & 61.2 & 41.5 & 25.6 & 22.2 & 54.6 & 49.7 & 25.4 & 19.9 & 47.0 & 18.1 & 26.0 & 38.9 & 67.7 & 2.0 & 22.6 & 41.1 & 34.3 & 37.9 & 55.3 & 37.9 \\
        PCL-OB-G VGG16~\cite{tang2018pcl} & 58.2 & \textbf{66.0} & 41.8 & 24.8 & \textbf{27.2} & 55.7 & \textbf{55.2} & 28.5 & 16.6 & \textbf{51.0} & 17.5 & 28.6 & \textbf{49.7} & \textbf{70.5} & 7.1 & \textbf{25.7} & 47.5 & 36.6 & 44.1 & \textbf{59.2} & 40.6 \\
        TS$^2$C~\cite{Wei_2018_ECCV} & 67.4 & 57.0 & 37.7 & 23.7 & 15.2 & \textbf{56.9} & 49.1 & 64.8 & 15.1 & 39.4 & 19.3 & \textbf{48.4} & 44.5 & 67.2 & 2.1 & 23.3 & 35.1 & 40.2 & \textbf{46.6} & 45.8 & 40.0 \\
        \hline
        OICR Ens.+FRCNN~\cite{Tang_2017_CVPR} & 71.4 & 69.4 & 55.1 & 29.8 & \textit{28.1} & 55.0 & \textit{57.9} & 24.4 & 17.2 & \textit{59.1} & 21.8 & 26.6 & 57.8 & 71.3 & 1.0 & 23.1 & \textit{52.7} & 37.5 & 33.5 & 56.6 & 42.5 \\
        PCL-OB-G Ens.+FRCNN~\cite{tang2018pcl} & 69.0 & \textit{71.3} & \textit{56.1} & 30.3 & 27.3 & 55.2 & 57.6 & 30.1 & 8.6 & 56.6 & 18.4 & 43.9 & \textit{64.6} & \textit{71.8} & 7.5 & 23.0 & 46.0 & 44.1 & 42.6 & 58.8 & 44.2 \\
        \hline
        \textbf{Ours} \rule{0pt}{1.01em} & \textbf{74.2} & 49.8 & \textbf{56.0} & \textbf{32.5} & 22.0 & 55.1 & 49.8 & \textbf{73.4} & \textbf{20.4} & 47.8 & \textbf{32.0} & 39.7 & 48.0 & 62.6 & \textbf{8.6} & 23.7 & \textbf{52.1} & \textbf{52.5} & 42.9 & 59.1 & \textbf{45.1} \\
    \Xhline{2\arrayrulewidth}
    \end{tabularx}
    \vspace{-0.3cm}
    \caption{\textbf{Average precision (in \%) on the Pascal VOC test set using image-level labels}. The top shows VOC 2007 and the bottom shows VOC 2012 results. The best single model is in \textbf{bold}, and best ensemble in \textit{italics}.} 
    \label{tab:result_voc07_map}
\vspace{-0.4cm}
\end{table*}

\vspace{-0.5cm}
\paragraph{Data v.s. performance.}

We show the potential of our model using Flickr30K and MIRFlickr1M~\cite{huiskes2010new}.
For the latter, we concatenate the title and all user-generated content tags to form caption annotation. We then use our text classifier learned on COCO to rule out examples unlikely to mention our target classes. 
This filtering results in a dataset with around 20\% of the original data, and we refer to it as Flickr200K.
We use 10\%, 20\%, 50\%, 100\% data from both datasets, and report average precision on VOC 2007. We see from Fig.~\ref{fig:data_vs_map} that as training data increases, mAP increases accordingly.
To estimate model potential, we fit a square root function to the rightmost four points in the figure and use it to estimate 54.4 mAP at 1 million samples.

\subsection{Using image labels as supervision}
\label{sec:results_using_label}

\begin{table}[t]
    \footnotesize
    \centering
    \begin{tabularx}{0.9\linewidth}{>{\hsize=2\hsize}X>{\hsize=0.5\hsize\centering\arraybackslash}X>{\hsize=0.5\hsize\centering\arraybackslash}X}
    \toprule
        \multirow{2}{*}{Methods} & \multicolumn{2}{c}{Avg. Precision, IoU} \\
        & 0.5:0.95 & 0.5 \\
    \midrule
        Faster RCNN~\cite{ren2015faster} & 21.9 & 42.7\\
        Faster Inception-V2~\cite{Huang_2017_CVPR} & 28.0 & - \\
    \midrule
        PCL-OB-G VGG16~\cite{tang2018pcl} & 8.5 & 19.4 \\
        PCL-OB-G Ens.+FRCNN~\cite{tang2018pcl} & 9.2 & 19.6 \\
        \textbf{Ours} & \textbf{10.6} & \textbf{23.4} \\
    \bottomrule
    \end{tabularx}
        \vspace{-0.2cm}
    \caption{\textbf{COCO detection using image-level labels}, with supervised detection models at the top, best WSOD in \textbf{bold}. }
    \label{tab:result_coco_map}
    \vspace{-0.2cm}
\end{table}

We finally show the performance of our method in the classic WSOD setting where \textit{image-level supervision} is available. These results validate the method component described in Sec.~\ref{sec:multiple_instance_detection}.
They also serve as an approximate \textit{upper bound} for the more challenging task in Sec.~\ref{sec:results_using_caption}.

\vspace{-0.5cm}
\paragraph{Results on PASCAL VOC.}
For each image, we extract object categories from all the ground-truth bounding boxes, and only keep these \emph{image-level} labels for training, discarding box information.
For VOC 2007 and 2012, we train on 5,011 and 11,540 \textit{trainval} images respectively and
evaluate on 4,952 and 10,991 \textit{test} images.\footnote{VOC 2012 result:  \url{http://host.robots.ox.ac.uk:8080/anonymous/NOR9IV.html}}
We report the standard mean Average Precision (mAP) at IoU $>0.5$.
We compare against multiple strong WSOD baselines.
The results are shown in Tab.~\ref{tab:result_voc07_map}, and our single model outperforms the baseline methods (sometimes even ensemble methods) by a large margin. On VOC 2007, our model improves the mAP of the state-of-the-art single method TS$^2$C method by 9\%. 
On VOC 2012, our method outperforms the strongest single-model baseline PCL-OB-G VGG16 by 11\%. Some prior work uses their WSOD detection results to further train an Fast-RCNN~\cite{Girshick_2015_ICCV} detector (denoted as ``+FRCNN'' in Tab. \ref{tab:result_voc07_map}) and obtain an additional 3 to 4 percents improvements on mAP. Even without such post-processing or ensemble, our model still achieves competitive performance on both VOC 2007 and 2012.

\begin{figure}[t]
    \centering
    \includegraphics[width=0.49\linewidth]{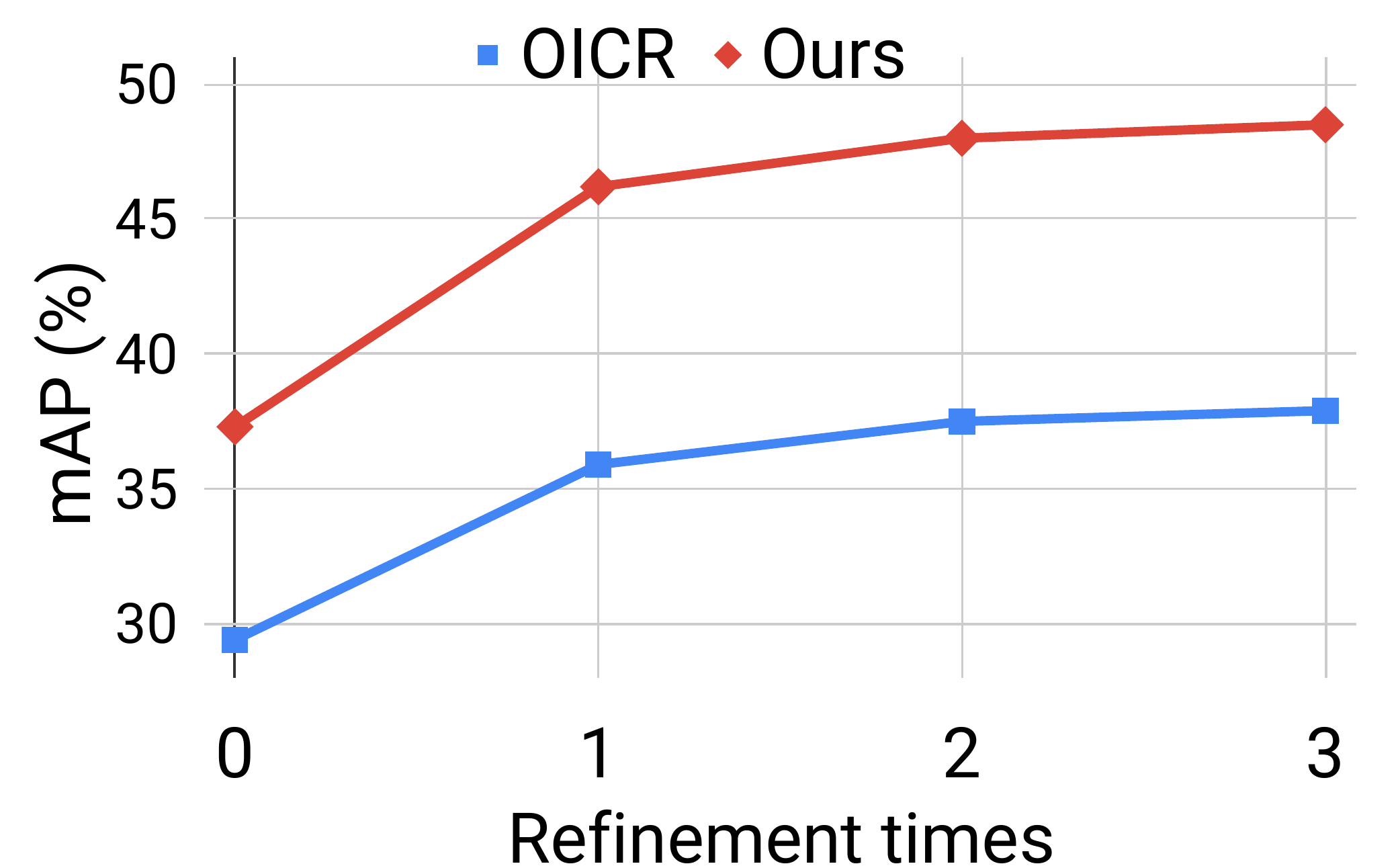}
    \includegraphics[width=0.49\linewidth]{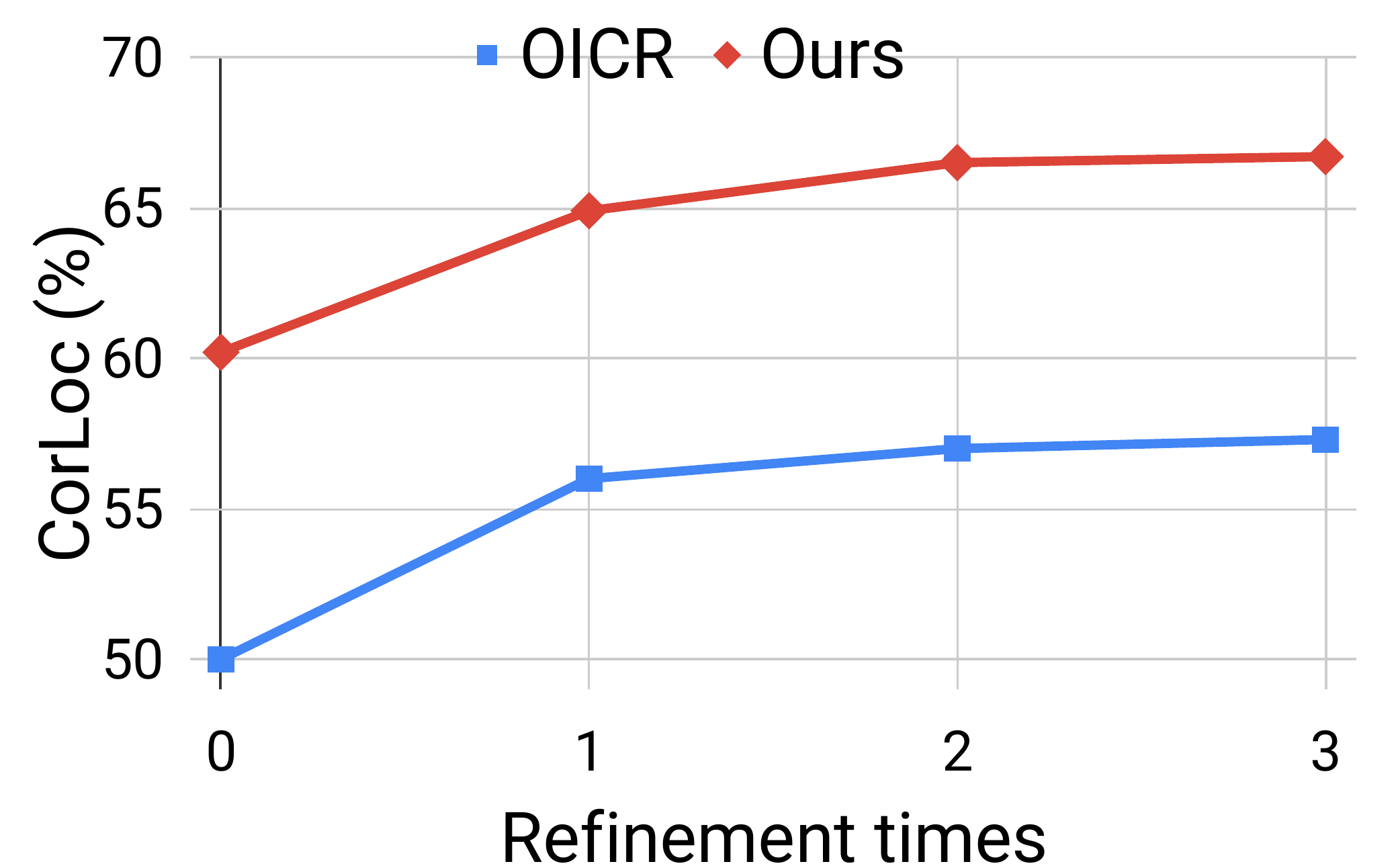}
    \vspace{-0.2cm}
    \caption{\textbf{Analysis of our basic network and OICR components on VOC 2007.} Comparison of the performance of our model and OICR VGG\_M after iterative refinement.}
    \label{fig:oicr_results}
\vspace{-0.4cm}
\end{figure}

\vspace{-0.5cm}
\paragraph{Effects of the basic network and OICR.}
The performance gain in our model comes from the following two aspects: (1) a more advanced detection model backbone architecture and (2) the online instance classifier refinement (OICR).
Fig.~\ref{fig:oicr_results} shows the performance of our method and that of Tang \etal~\cite{Tang_2017_CVPR} (\textsc{OICR VGG\_M}), both refining for 0, 1, 2, 3 times. With no (0) refinement, 
our basic network architecture outperforms the VGG\_M backbone of Tang \textit{et al.} by 27\% in mAP.
But the basic architecture improvement is not sufficient to achieve top results.
If we use OICR to refine the models 1, 2, or 3 times, we gain 24\%, 29\%, and 30\% respectively while Tang \etal achieve smaller improvement (22\%, 28\%, and 29\% gains). 

\vspace{-0.5cm}
\paragraph{Results on COCO.} 
We train our model on the 118,287 \textit{train2017} images, using the image-level ground truth labels. We report mAP at $\text{IoU=.50:.05:.95}$ and mAP@0.5, on the 20,288 \textit{test-dev2017} images.  
We compare to a representative fully-supervised detection model~\cite{ren2015faster}; ``Faster Inception-V2'' \cite{Huang_2017_CVPR} which is our method's supervised detection counterpart, 
and a recent WSOD model~\cite{tang2018pcl}. As demonstrated in Tab.~\ref{tab:result_coco_map}, our model outperforms the previous state-of-the-art WSOD method (PCL-OB-G Ens + FRCNN) by 15\% in terms of mAP, but the gap between general WSOD methods (including ours) and the supervised methods is still large due to the disparate supervision strength.
\section{Conclusion}
\label{sec:conclusion}

We showed how we can successfully leverage naturally arising, weak supervision in the form of captions. We amplify the signal that captions provide by learning to bridge the gap between what human annotators mention, and what is present in the image.
In the future, we will further verify the benefit of our approach by training on a larger, but even noisier dataset, namely WebVision~\cite{li2017webvision}. We will also extend our method to incorporate raw supervision in the form of spoken descriptions in video. Finally, we will experiment with incorporating the text classifier training and the weakly supervised detection into the same end-to-end framework.

\vspace{-0.5cm}
\paragraph{Acknowledgement:} 
This  material  is  based  upon  work supported by the National Science Foundation under Grant No. 1566270. This research was also supported by Google Faculty  Research  Awards and  an NVIDIA hardware grant. 

{\small
\bibliographystyle{ieee_fullname}
\bibliography{egbib}

\begin{thebibliography}{10}\itemsep=-1pt

\bibitem{Abadi:2016:TSL:3026877.3026899}
Mart\'{\i}n Abadi, Paul Barham, Jianmin Chen, Zhifeng Chen, Andy Davis, Jeffrey
  Dean, Matthieu Devin, Sanjay Ghemawat, Geoffrey Irving, Michael Isard,
  Manjunath Kudlur, Josh Levenberg, Rajat Monga, Sherry Moore, Derek~G. Murray,
  Benoit Steiner, Paul Tucker, Vijay Vasudevan, Pete Warden, Martin Wicke, Yuan
  Yu, and Xiaoqiang Zheng.
\newblock Tensorflow: A system for large-scale machine learning.
\newblock In {\em Proceedings of the 12th USENIX Conference on Operating
  Systems Design and Implementation}, OSDI'16, pages 265--283, Berkeley, CA,
  USA, 2016. USENIX Association.

\bibitem{Anderson_2018_CVPR}
Peter Anderson, Xiaodong He, Chris Buehler, Damien Teney, Mark Johnson, Stephen
  Gould, and Lei Zhang.
\newblock Bottom-up and top-down attention for image captioning and visual
  question answering.
\newblock In {\em Proceedings of the IEEE Conference on Computer Vision and
  Pattern Recognition (CVPR)}, June 2018.

\bibitem{berg2012understanding}
Alexander~C Berg, Tamara~L Berg, Hal Daume, Jesse Dodge, Amit Goyal, Xufeng
  Han, Alyssa Mensch, Margaret Mitchell, Aneesh Sood, Karl Stratos, et~al.
\newblock Understanding and predicting importance in images.
\newblock In {\em Proceedings of the IEEE Conference on Computer Vision and
  Pattern Recognition (CVPR)}, June 2012.

\bibitem{Bilen_2016_CVPR}
Hakan Bilen and Andrea Vedaldi.
\newblock Weakly supervised deep detection networks.
\newblock In {\em Proceedings of the IEEE Conference on Computer Vision and
  Pattern Recognition (CVPR)}, June 2016.

\bibitem{opencv_library}
Gary Bradski.
\newblock The opencv library.
\newblock {\em Dr. Dobb's Journal of Software Tools}, 2000.

\bibitem{Chen_2017_CVPR}
Kai Chen, Hang Song, Chen Change~Loy, and Dahua Lin.
\newblock Discover and learn new objects from documentaries.
\newblock In {\em Proceedings of the IEEE Conference on Computer Vision and
  Pattern Recognition (CVPR)}, July 2017.

\bibitem{chen2015microsoft}
Xinlei Chen, Hao Fang, Tsung-Yi Lin, Ramakrishna Vedantam, Saurabh Gupta, Piotr
  Doll{\'a}r, and C~Lawrence Zitnick.
\newblock Microsoft coco captions: Data collection and evaluation server.
\newblock {\em arXiv preprint arXiv:1504.00325}, 2015.

\bibitem{chen2018domain}
Yuhua Chen, Wen Li, Christos Sakaridis, Dengxin Dai, and Luc Van~Gool.
\newblock Domain adaptive faster r-cnn for object detection in the wild.
\newblock In {\em Proceedings of the IEEE Conference on Computer Vision and
  Pattern Recognition (CVPR)}, June 2018.

\bibitem{imagenet_cvpr09}
Jia Deng, Wei Dong, Richard Socher, Li-Jia Li, Kai Li, and Li Fei-Fei.
\newblock Imagenet: A large-scale hierarchical image database.
\newblock In {\em Proceedings of the IEEE Conference on Computer Vision and
  Pattern Recognition (CVPR)}, June 2009.

\bibitem{Diba_2017_CVPR}
Ali Diba, Vivek Sharma, Ali Pazandeh, Hamed Pirsiavash, and Luc Van~Gool.
\newblock Weakly supervised cascaded convolutional networks.
\newblock In {\em Proceedings of the IEEE Conference on Computer Vision and
  Pattern Recognition (CVPR)}, July 2017.

\bibitem{Girshick_2015_ICCV}
Ross Girshick.
\newblock Fast r-cnn.
\newblock In {\em Proceedings of the IEEE International Conference on Computer
  Vision (ICCV)}, December 2015.

\bibitem{gopalan2011domain}
Raghuraman Gopalan, Ruonan Li, and Rama Chellappa.
\newblock Domain adaptation for object recognition: An unsupervised approach.
\newblock In {\em Proceedings of the IEEE International Conference on Computer
  Vision (ICCV)}, November 2011.

\bibitem{He_2017_ICCV}
Kaiming He, Georgia Gkioxari, Piotr Dollar, and Ross Girshick.
\newblock Mask r-cnn.
\newblock In {\em Proceedings of the IEEE International Conference on Computer
  Vision (ICCV)}, October 2017.

\bibitem{hu2016segmentation}
Ronghang Hu, Marcus Rohrbach, and Trevor Darrell.
\newblock Segmentation from natural language expressions.
\newblock In {\em Proceedings of the European Conference on Computer Vision
  (ECCV)}, October 2016.

\bibitem{Huang_2017_CVPR}
Jonathan Huang, Vivek Rathod, Chen Sun, Menglong Zhu, Anoop Korattikara,
  Alireza Fathi, Ian Fischer, Zbigniew Wojna, Yang Song, Sergio Guadarrama, and
  Kevin Murphy.
\newblock Speed/accuracy trade-offs for modern convolutional object detectors.
\newblock In {\em Proceedings of the IEEE Conference on Computer Vision and
  Pattern Recognition (CVPR)}, July 2017.

\bibitem{huiskes2010new}
Mark~J Huiskes, Bart Thomee, and Michael~S Lew.
\newblock New trends and ideas in visual concept detection: the mir flickr
  retrieval evaluation initiative.
\newblock In {\em Proceedings of the international conference on Multimedia
  information retrieval}, pages 527--536. ACM, 2010.

\bibitem{kantorov2016contextlocnet}
Vadim Kantorov, Maxime Oquab, Minsu Cho, and Ivan Laptev.
\newblock Contextlocnet: Context-aware deep network models for weakly
  supervised localization.
\newblock In {\em Proceedings of the European Conference on Computer Vision
  (ECCV)}, October 2016.

\bibitem{Krause_2017_CVPR}
Jonathan Krause, Justin Johnson, Ranjay Krishna, and Li Fei-Fei.
\newblock A hierarchical approach for generating descriptive image paragraphs.
\newblock In {\em Proceedings of the IEEE Conference on Computer Vision and
  Pattern Recognition (CVPR)}, July 2017.

\bibitem{li2016weakly}
Dong Li, Jia-Bin Huang, Yali Li, Shengjin Wang, and Ming-Hsuan Yang.
\newblock Weakly supervised object localization with progressive domain
  adaptation.
\newblock In {\em Proceedings of the IEEE Conference on Computer Vision and
  Pattern Recognition (CVPR)}, June 2016.

\bibitem{li2017webvision}
Wen Li, Limin Wang, Wei Li, Eirikur Agustsson, and Luc Van~Gool.
\newblock Webvision database: Visual learning and understanding from web data.
\newblock {\em arXiv preprint arXiv:1708.02862}, 2017.

\bibitem{Lu_2018_CVPR}
Jiasen Lu, Jianwei Yang, Dhruv Batra, and Devi Parikh.
\newblock Neural baby talk.
\newblock In {\em Proceedings of the IEEE Conference on Computer Vision and
  Pattern Recognition (CVPR)}, June 2018.

\bibitem{Misra_2016_CVPR}
Ishan Misra, C. Lawrence~Zitnick, Margaret Mitchell, and Ross Girshick.
\newblock Seeing through the human reporting bias: Visual classifiers from
  noisy human-centric labels.
\newblock In {\em Proceedings of the IEEE Conference on Computer Vision and
  Pattern Recognition (CVPR)}, June 2016.

\bibitem{Oquab_2015_CVPR}
Maxime Oquab, Leon Bottou, Ivan Laptev, and Josef Sivic.
\newblock Is object localization for free? - weakly-supervised learning with
  convolutional neural networks.
\newblock In {\em Proceedings of the IEEE Conference on Computer Vision and
  Pattern Recognition (CVPR)}, June 2015.

\bibitem{pennington2014glove}
Jeffrey Pennington, Richard Socher, and Christopher Manning.
\newblock Glove: Global vectors for word representation.
\newblock In {\em Proceedings of the 2014 conference on empirical methods in
  natural language processing (EMNLP)}, pages 1532--1543, 2014.

\bibitem{redmon2016you}
Joseph Redmon, Santosh Divvala, Ross Girshick, and Ali Farhadi.
\newblock You only look once: Unified, real-time object detection.
\newblock In {\em Proceedings of the IEEE Conference on Computer Vision and
  Pattern Recognition (CVPR)}, June 2016.

\bibitem{ren2015faster}
Shaoqing Ren, Kaiming He, Ross Girshick, and Jian Sun.
\newblock Faster r-cnn: Towards real-time object detection with region proposal
  networks.
\newblock In {\em Advances in neural information processing systems}, pages
  91--99, 2015.

\bibitem{rohrbach2016grounding}
Anna Rohrbach, Marcus Rohrbach, Ronghang Hu, Trevor Darrell, and Bernt Schiele.
\newblock Grounding of textual phrases in images by reconstruction.
\newblock In {\em Proceedings of the European Conference on Computer Vision
  (ECCV)}, October 2016.

\bibitem{Szegedy_2016_CVPR}
Christian Szegedy, Vincent Vanhoucke, Sergey Ioffe, Jon Shlens, and Zbigniew
  Wojna.
\newblock Rethinking the inception architecture for computer vision.
\newblock In {\em Proceedings of the IEEE Conference on Computer Vision and
  Pattern Recognition (CVPR)}, June 2016.

\bibitem{tang2012shifting}
Kevin Tang, Vignesh Ramanathan, Li Fei-Fei, and Daphne Koller.
\newblock Shifting weights: Adapting object detectors from image to video.
\newblock In {\em Advances in Neural Information Processing Systems}, pages
  638--646, 2012.

\bibitem{tang2018pcl}
Peng Tang, Xinggang Wang, Song Bai, Wei Shen, Xiang Bai, Wenyu Liu, and
  Alan~Loddon Yuille.
\newblock Pcl: Proposal cluster learning for weakly supervised object
  detection.
\newblock {\em IEEE transactions on pattern analysis and machine intelligence},
  2018.

\bibitem{Tang_2017_CVPR}
Peng Tang, Xinggang Wang, Xiang Bai, and Wenyu Liu.
\newblock Multiple instance detection network with online instance classifier
  refinement.
\newblock In {\em Proceedings of the IEEE Conference on Computer Vision and
  Pattern Recognition (CVPR)}, July 2017.

\bibitem{Teney_2018_CVPR}
Damien Teney, Peter Anderson, Xiaodong He, and Anton van~den Hengel.
\newblock Tips and tricks for visual question answering: Learnings from the
  2017 challenge.
\newblock In {\em Proceedings of the IEEE Conference on Computer Vision and
  Pattern Recognition (CVPR)}, June 2018.

\bibitem{thomas2018artistic}
Christopher Thomas and Adriana Kovashka.
\newblock Artistic object recognition by unsupervised style adaptation.
\newblock In {\em Asian Conference on Computer Vision (ACCV)}, pages 460--476.
  Springer, 2018.

\bibitem{Turakhia_2013_ICCV}
Naman Turakhia and Devi Parikh.
\newblock Attribute dominance: What pops out?
\newblock In {\em Proceedings of the IEEE International Conference on Computer
  Vision (ICCV)}, December 2013.

\bibitem{uijlings2013selective}
Jasper~RR Uijlings, Koen~EA Van De~Sande, Theo Gevers, and Arnold~WM Smeulders.
\newblock Selective search for object recognition.
\newblock {\em International Journal of Computer Vision}, 104(2):154--171,
  2013.

\bibitem{Wan_2018_CVPR}
Fang Wan, Pengxu Wei, Jianbin Jiao, Zhenjun Han, and Qixiang Ye.
\newblock Min-entropy latent model for weakly supervised object detection.
\newblock In {\em Proceedings of the IEEE Conference on Computer Vision and
  Pattern Recognition (CVPR)}, June 2018.

\bibitem{Wei_2018_ECCV}
Yunchao Wei, Zhiqiang Shen, Bowen Cheng, Honghui Shi, Jinjun Xiong, Jiashi
  Feng, and Thomas Huang.
\newblock Ts2c: Tight box mining with surrounding segmentation context for
  weakly supervised object detection.
\newblock In {\em Proceedings of the European Conference on Computer Vision
  (ECCV)}, September 2018.

\bibitem{Ye_2018_ECCV}
Keren Ye and Adriana Kovashka.
\newblock Advise: Symbolism and external knowledge for decoding advertisements.
\newblock In {\em Proceedings of the European Conference on Computer Vision
  (ECCV)}, September 2018.

\bibitem{young2014image}
Peter Young, Alice Lai, Micah Hodosh, and Julia Hockenmaier.
\newblock From image descriptions to visual denotations: New similarity metrics
  for semantic inference over event descriptions.
\newblock {\em Transactions of the Association for Computational Linguistics},
  2:67--78, 2014.

\bibitem{Zhang_2018_BMVC}
Mingda Zhang, Rebecca Hwa, and Adriana Kovashka.
\newblock Equal but not the same: Understanding the implicit relationship
  between persuasive images and text.
\newblock In {\em Proceedings of the British Machine Vision Conference (BMVC)},
  September 2018.

\bibitem{Zhou_2016_CVPR}
Bolei Zhou, Aditya Khosla, Agata Lapedriza, Aude Oliva, and Antonio Torralba.
\newblock Learning deep features for discriminative localization.
\newblock In {\em Proceedings of the IEEE Conference on Computer Vision and
  Pattern Recognition (CVPR)}, June 2016.

\bibitem{zitnick2014edge}
C~Lawrence Zitnick and Piotr Doll{\'a}r.
\newblock Edge boxes: Locating object proposals from edges.
\newblock In {\em Proceedings of the European Conference on Computer Vision
  (ECCV)}, September 2014.

\end{thebibliography}
}

\onecolumn

\setcounter{section}{0}
\setcounter{figure}{0}
\setcounter{table}{0}
\setcounter{footnote}{0}
\emptythanks

\title{Cap2Det: Learning to Amplify Weak Caption Supervision for Object Detection
(Supplementary file)}

\makeatletter
\renewcommand{\AB@affillist}{}
\renewcommand{\AB@authlist}{}
\setcounter{authors}{0}
\makeatother

\author{\vspace{-3cm}}

\maketitle
\thispagestyle{empty}

In this document, we include more information and statistics about our Cap2Det model. We also provide additional quantitative and qualitative experimental results.

We first show more statistics about our Cap2Det model. We provide per-class precision and recall of our label inference module (Sec. 3.1 in the paper) in Sec. 1 to help better understand how the module affects the final detection performance. We find that some of the categories benefit from the module more than others. We also qualitatively analyze two strong baseline methods in Sec. 2. 

Next, we provide additional details of our implementation of the Online Instance Classifier Refinement (OICR) module (Sec. 3.2.2 in the paper) in Sec. 3. Generally speaking, ours is a simplified version without re-weighing the OICR loss as compared to Tang \etal~[31].

We show more statistics about our learned models. For our model learned from COCO captions (Sec. 4.2 in the paper), we show more metrics computed by the COCO evaluation server in Sec. 4. In Sec. 5, we measure our model learned from Pascal VOC labels (Sec. 4.4 in the paper), using the Correct Localization scores. This evaluation is also proceeded in [30,31,36].

Finally, to enable a more intuitive understanding of our model, we provide more qualitative results of the models learned from both COCO and Flickr30K (Sec. 4.2 in our paper). For the model learned from COCO (Sec. 6), we side-by-side compare our \textsc{EM+TextClsf} method to the \textsc{ExactMatch} baseline. For the model learned from Flickr30K (Sec. 7), we, in addition, show the predicted image-level pseudo label. Both results explain the benefits of our proposed idea of amplifying weak caption supervision.

\newpage
\section{Analysis of per-class precision/recall of label inference method}

We provide per-class precision/recall of our label inference method (Sec. 3.1) to see how this method affects the detection performance. We still use the 5,000 COCO \textit{val} examples, but evaluate on only the overlapped classes (20 classes) between COCO and Pascal. We show the comparison between our label inference method \textsc{EM+TextClsf} and the lexical matching method \textsc{ExactMatch}.

Fig.1 shows the comparison. Our method does not affect the precision too much, but it has a positive impact on the recall. 
As compared to the Tab. 1 of our paper, not every percentage of improvements of the recall in the text inference leads to an increase of performance in object detection.
However, there are some notable classes that can be explained by the merits of text inference model. For example, using our method, the the recall rate of ``cow'', ``horse'', ``person'', are increased by 47.2\%, 26.0\%, 54.2\% respectively (cow from 53\% to 78\%, horse from 73\% to 92\%, person from 24\% to 37\%). Their detection mAP, accordingly, are increased by 24.9\%, 16.5\%, 62.5\% (cow from 49.0\% to 61.2\%, horse from 44.2\% to 51.5\%, person from 10.4\% to 16.9\%).

\begin{figure}[ht]
    \centering
    \includegraphics[width=0.9\linewidth]{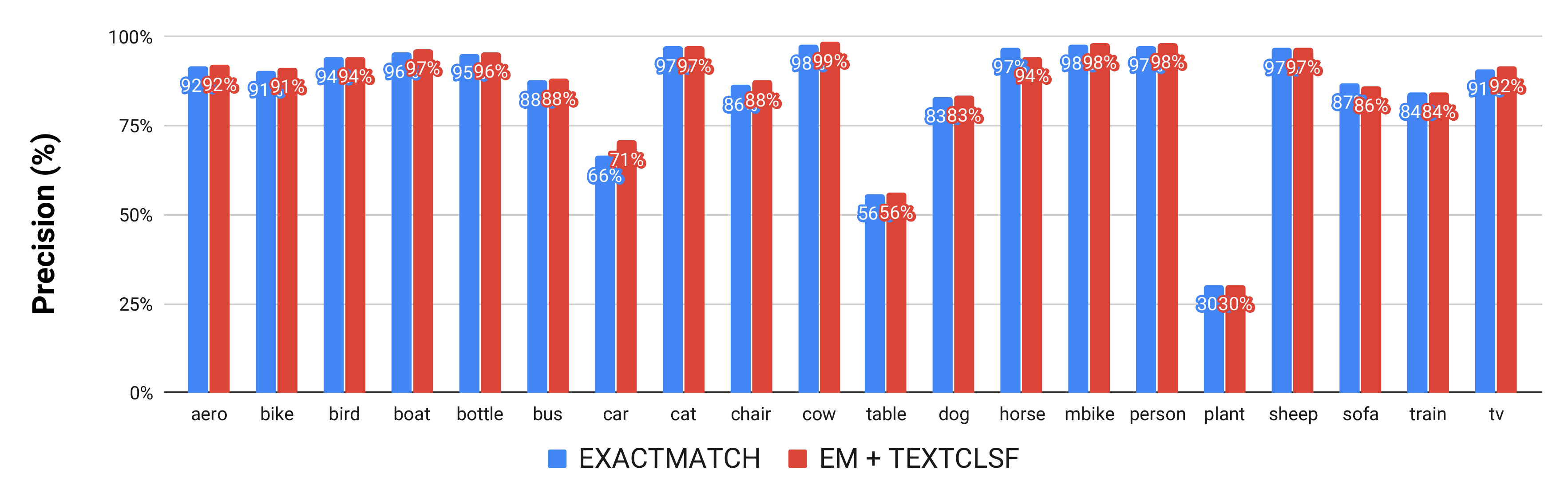}
    \includegraphics[width=0.9\linewidth]{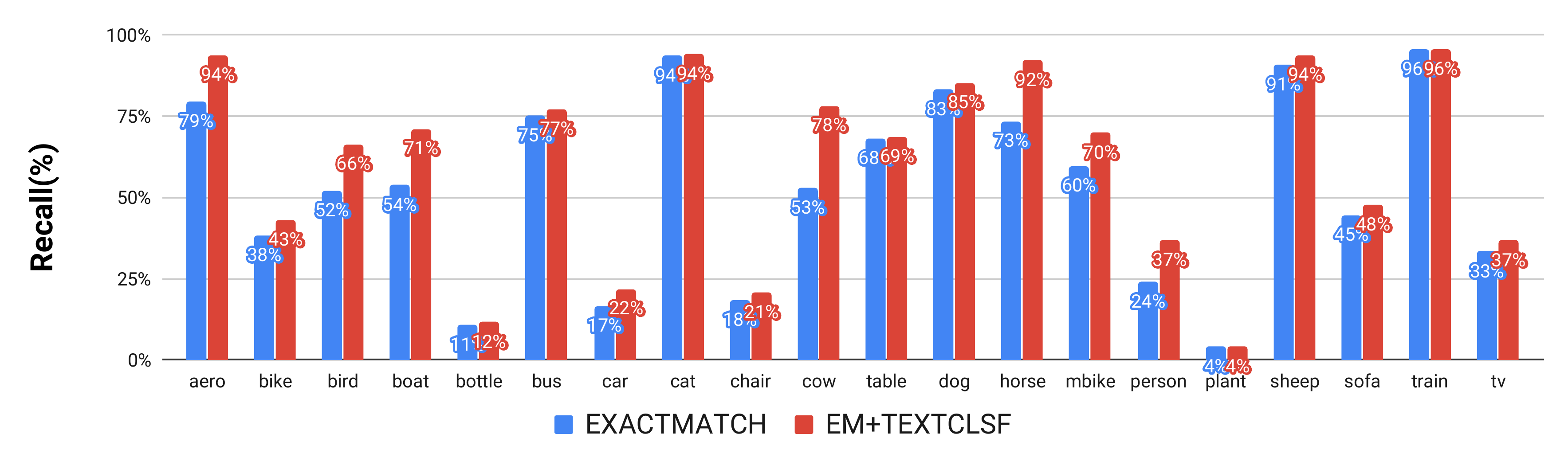}
    \caption{\textbf{Precision/recall of the Pascal labels.} Similar to the Fig. 3 in our paper we evaluate the precision and recall, but we focus on the subset of the 20 Pascal VOC classes instead of the performance on the 80 COCO labels.}
    \label{fig:per_class_text_pr}
\end{figure}

\newpage
\section{Qualitative analysis of the word embedding based methods}

We show some qualitative analysis of the \textsc{GloVePseudo} and \textsc{LearnedGloVe}, by visualizing their word embedding feature space. Intuitively,  Fig.2 shows that the Glove embedding optimized on the general purpose textual corpus is unable to distinguish the nuance such as bicycle and motorcycle, pizza and sandwich, etc. This explains the improved performance of \textsc{LearnedGloVe} compared to \textsc{GloVePseudo} (Tab. 1 in the paper).

\begin{figure}[ht]
    \centering
    \begin{subfigure}[b]{0.4\textwidth}
        \includegraphics[width=1.0\textwidth]{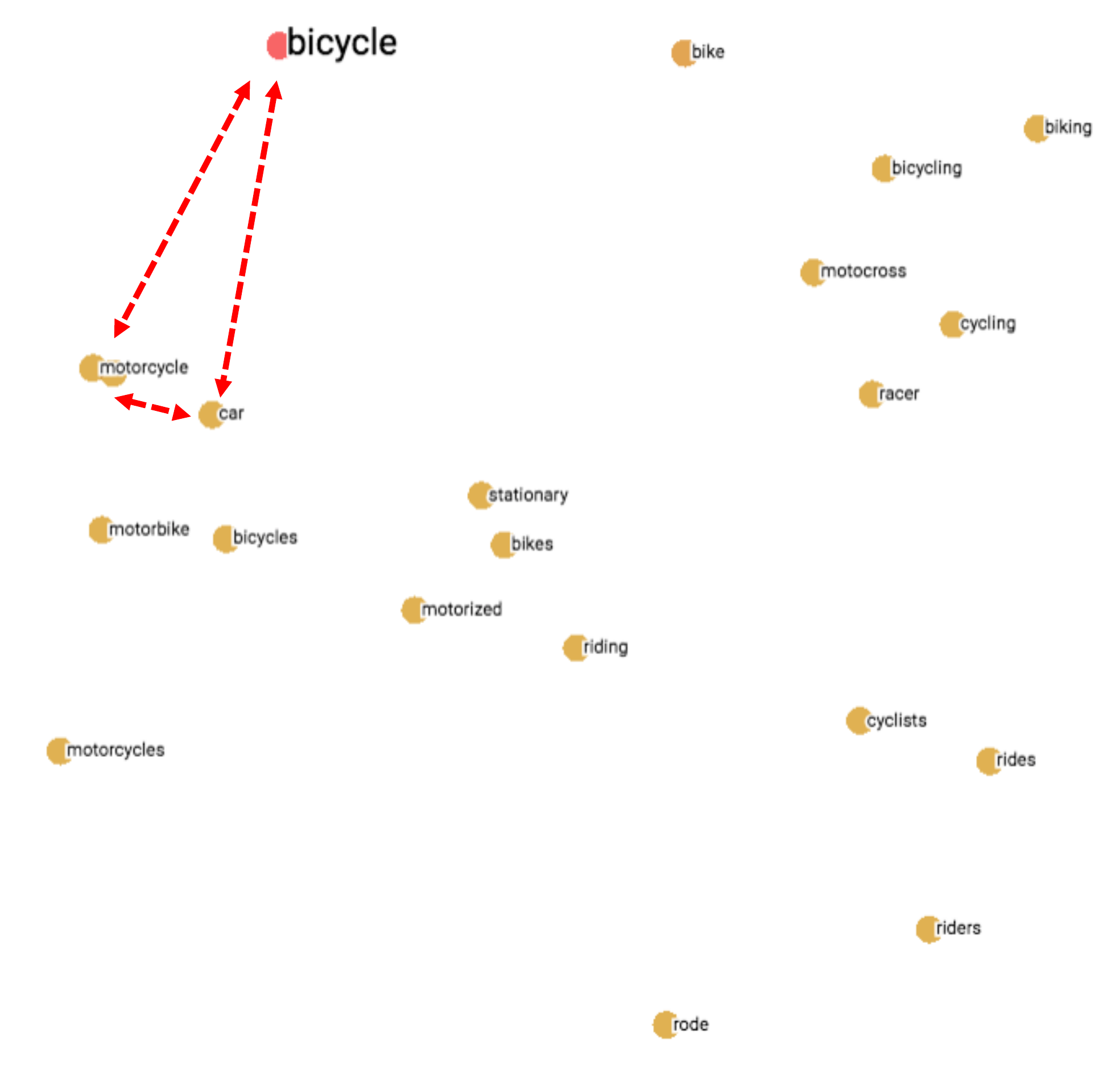}
        \caption{``bicycle'' in Glove space: visually different objects such as ``car'' and ``motorcycle'' are near to ``bicycle''.}
        \label{fig:glove_bicycle}
    \end{subfigure}
    ~~~~~~~~
    \begin{subfigure}[b]{0.4\textwidth}
        \includegraphics[width=1.0\textwidth]{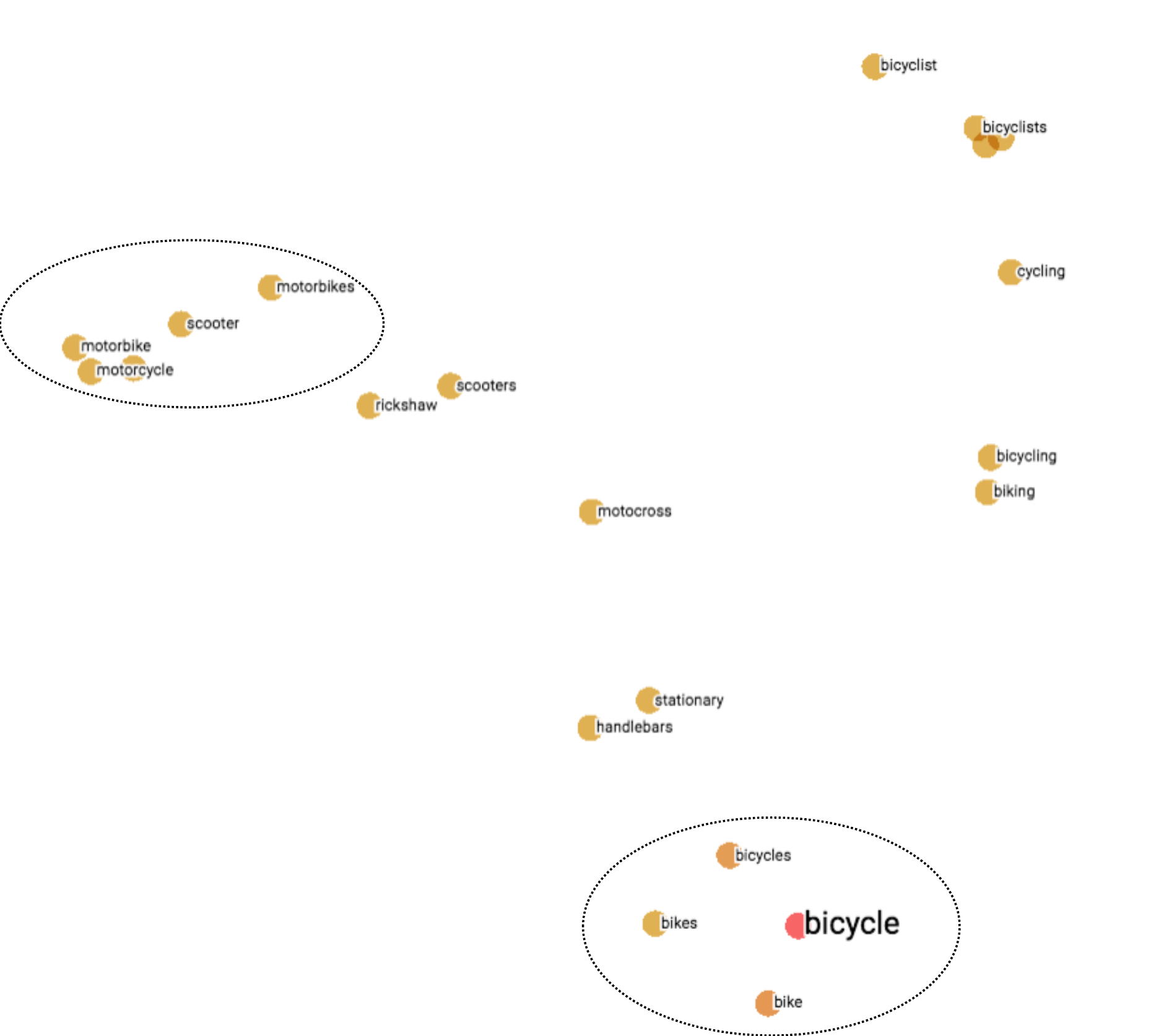}
        \caption{``bicycle'' in Learned Glove space: visually similar objects are clustered.}
        \label{fig:learned_glove_bicycle}
    \end{subfigure}
    \begin{subfigure}[b]{0.4\textwidth}
        \includegraphics[width=1.0\textwidth]{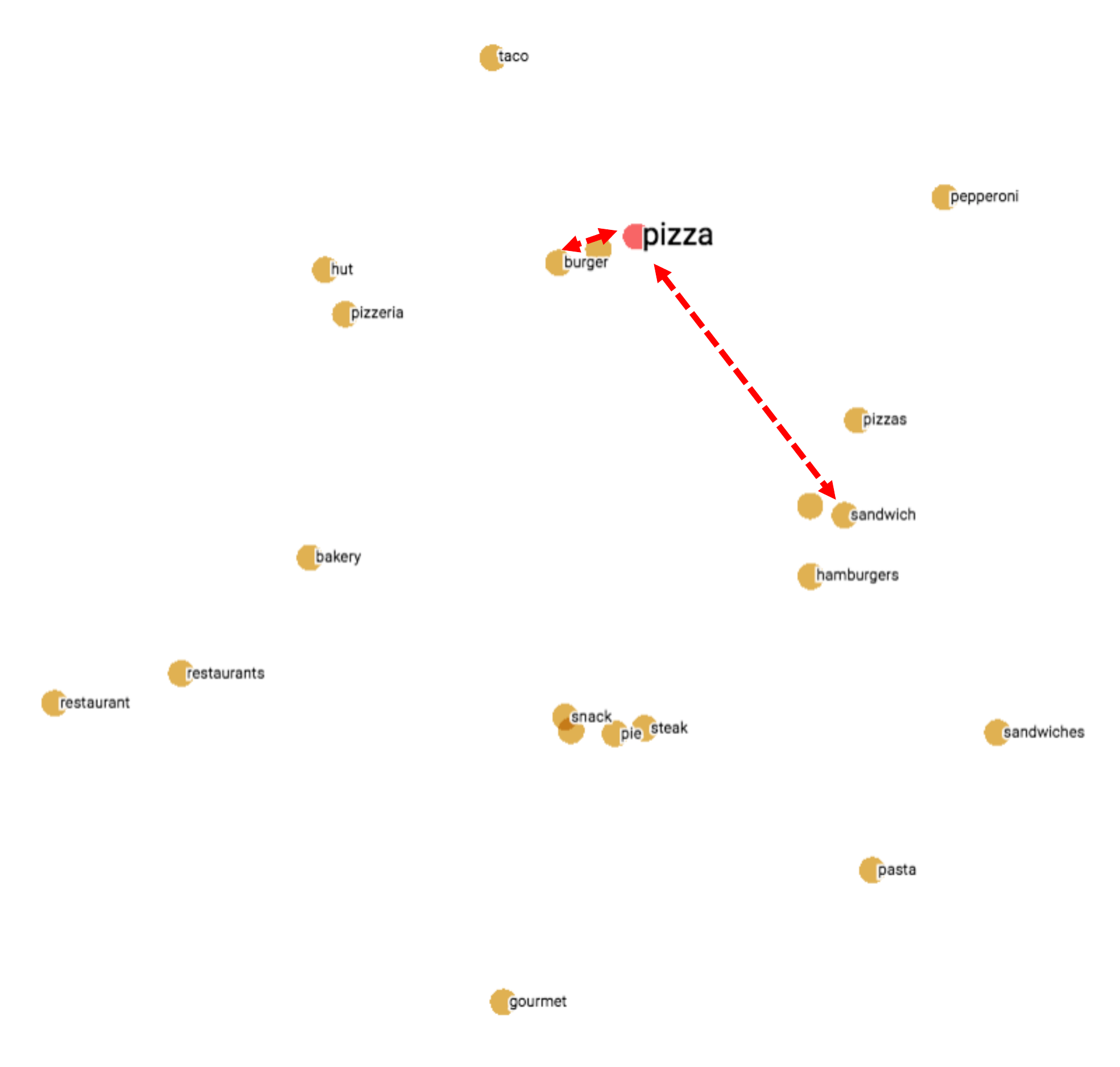}
        \caption{``pizza'' in Glove space: visually different object ``burger'' is almost indistinguishable.}
        \label{fig:glove_pizza}
    \end{subfigure}
    ~~~~~~~~
    \begin{subfigure}[b]{0.4\textwidth}
        \includegraphics[width=1.0\textwidth]{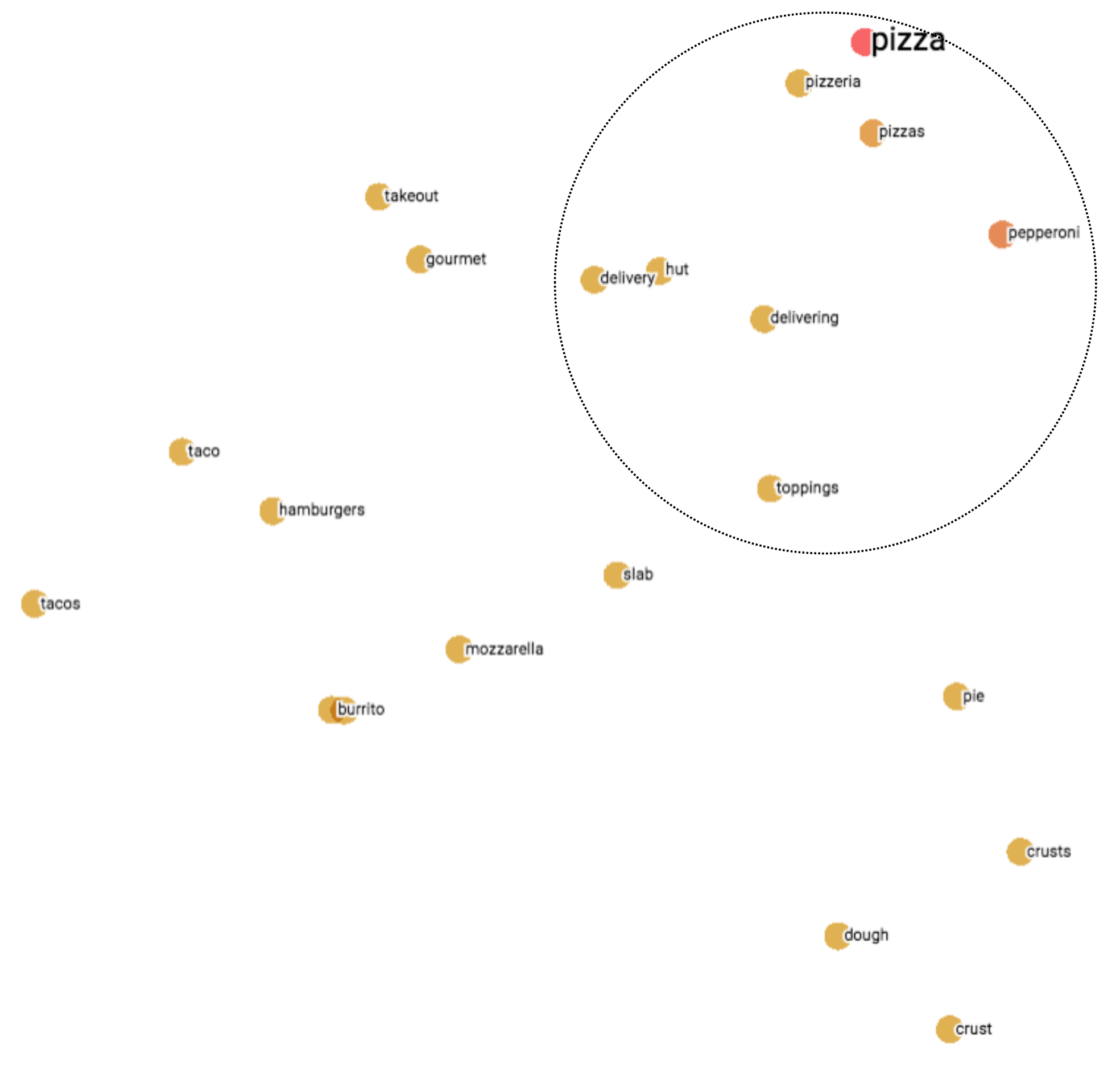}
        \caption{``pizza'' in Learned Glove space: ``pizza'' related words such as ``pepperoni'', ``toppings'' are clustered.}
        \label{fig:learned_glove_pizza}
    \end{subfigure}
    \caption{\textbf{Visualization of the embedding space for two strong baselines \textsc{GloVePseudo} and \textsc{LearnedGloVe}}. We show the 20 nearest neighbors to the query word ``bicycle'' and ``pizza''. Both (a) and (c) visualize the original Glove feature space while (b) and (d) visualize the learned Glove embedding.}
    \label{fig:by}
\end{figure}

\newpage
\section{Online instance classifier refinement}

We provide more details regarding our Sec. 3.2.2, namely, the Online Instance Classifier Refinement (OICR). Generally speaking, the instance-level label $\hat{\bm{y}}^{(k+1)}$ at the $(k+1)$-th iteration is inferred from both the image level label $[y_1,\dots,y_C]$ and the detection score $\hat{\bm{s}}^{(k)}$ at $k$-th iteration using Algorithm 1. Then, it is used to guide the learning of the  $\hat{\bm{s}}^{(k+1)}$ using Eq.5 in the paper.

\begin{algorithm}
\caption{Online Instance Classifier Refinement - Generating Pseudo Instance Level Labels at the $(k+1)$-th Iteration}
\label{alg:oicr}
\SetKwInOut{Input}{Input}
\SetKwInOut{Output}{Output}
\Input{Proposals $[r_1,\dots,r_m]$; \\
Image level labels $[y_1,\dots,y_C]$; \\
Detection scores at the $k$-th iteration $\hat{\bm{s}}^{(k)}=[s_{1,1}^{(k)},\dots,s_{m,C+1}^{(k)}]$; \\
Iou threshold $threshold$.}
\Output{Instance level labels at the $(k+1)$-th iteration $\hat{\bm{y}}^{(k+1)}=[\hat{y}_{1,1}^{(k+1)},\dots,\hat{y}_{m,(C+1)}^{(k+1)}]$.}

$\hat{\bm{y}}^{(k+1)} \gets \vec{\bm{0}}$; \\
\For{$c' \gets 1$ \textbf{to} $C$} {
    \If{$y_{c'} = 1$} {
        $j \gets \text{arg}\max\limits_i s_{i,c'}^{(k)}$; \\
        \For{$i \gets 1$ \textbf{to} $m$} {
            \If{$\text{IoU}(r_i, r_j) > \text{threshold}$} {
                $\hat{y}_{i,c'}^{(k+1)} \gets 1$; \footnotesize\ttfamily\textcolor{blue}{// Assign foreground target.}
            }
        }
    }
}
\For{$i \gets 1$ \textbf{to} $m$} {
    $t \gets \sum\limits_c \hat{y}_{i,c}^{(k+1)}$; \\
    \If{$t=0$}{ 
        $t \gets 1$; \\
        $\hat{y}_{i,C+1}^{(k+1)} \gets 1$; \footnotesize\ttfamily\textcolor{blue}{// Assign background target.}
    }
    \For{$c \gets 1$ \textbf{to} $C$} {
        $\hat{y}_{i, c}^{(k+1)} \gets \hat{y}_{i, c}^{(k+1)} / t$; \footnotesize\ttfamily\textcolor{blue}{// Make the probabilities sum to 1.}
    }
}
\Return $\hat{\bm{y}}^{(k+1)}$
\end{algorithm}

\newpage
\section{Using captions as supervision}

We provide more metrics to measure our model on the COCO dataset, including the Average Recall (AR) given different numbers of detected boxes (max=1,10,100), and the Average Recall Across Scales (size=small, median, large). Please check the COCO object detection challenge to see the details of these standard metrics. The observations are still similar to the discussion of the Sec. 4.2 of our paper. The following table shows the result.

\begin{table*}[ht]
    \footnotesize
    \centering
    \setlength\tabcolsep{0pt} 
    \begin{tabularx}{\linewidth}{>{\hsize=2.5\hsize}X|*{3}{>{\hsize=0.875\hsize\centering\arraybackslash}X}|*{3}{>{\hsize=0.875\hsize\centering\arraybackslash}X}|*{3}{>{\hsize=0.875\hsize\centering\arraybackslash}X}|*{3}{>{\hsize=0.875\hsize\centering\arraybackslash}X}}
    \toprule
        \multirow{2}{*}{Methods} & \multicolumn{3}{c|}{Avg. Precision, IoU} & \multicolumn{3}{c}{Avg. Precision, Area} & \multicolumn{3}{c}{Avg. Recall, \#Dets} & \multicolumn{3}{c}{Avg. Recall, Area} \\
        & 0.5:0.95 & 0.5 & 0.75 & S & M & L & 1 & 10 & 100 & S & M & L \\
    \midrule
        \textsc{GT-Label} & 10.6 & 23.4 & 8.7 & 3.2 & 12.1 & 18.1 & 13.6 & 20.9 & 21.4 & 4.5 & 23.1 & 39.3 \\
    \hline
        \textsc{ExactMatch (EM)} \rule{0px}{1.01em} & 8.9 & 19.7 & 7.1 & 2.3 & 10.1 & 16.3 & \textbf{12.6} & \textbf{19.3} & \textbf{19.8} & 3.4 & 20.3 & 37.4 \\
        \textsc{EM + GloVePseudo} & 8.6 & 19.0 & 6.9 & 2.2 & 10.0 & 16.0 & 12.2 & 18.7 & 18.9 & 2.9 & 19.0 & 37.6 \\
        \textsc{EM + LearnedGloVe} & 8.9 & 19.7 & 7.2 & 2.5 & 10.4 & \textbf{16.6} & 12.3 & 19.1 & 19.6 & \textbf{3.5} & 20.0 & 37.7 \\
        \textsc{EM + ExtendVocab} & 8.8 & 19.4 & 7.1 & 2.3 & 10.5 & 16.1 & 12.1 & 19.0 & 19.5 & 3.4 & 20.3 & 37.5 \\
        \textsc{EM + TextClsf}& \textbf{9.1} & \textbf{20.2} & \textbf{7.3} & \textbf{2.6} & \textbf{10.8} & \textbf{16.6} & 12.5 & \textbf{19.3} & \textbf{19.8} & \textbf{3.5} & \textbf{20.6} & \textbf{37.8} \\
    \bottomrule
    \end{tabularx}
    \caption{\textbf{COCO test-dev results (learning from COCO captions)}. We report these numbers by submitting to the COCO evaluation server. The best method is shown in \textbf{bold}.}
    \label{tab:result_cap_coco_map_full}
\end{table*}

\section{Using image labels as supervision}

Similar to [30,31,36], we also report the Correct Localization (CorLoc) scores (in \%) of our method, using the Pascal VOC \textit{trainval} set. We employ the same threshold of IoU ($>=0.5$) as that in Tab.3 of our paper. The results are shown in the following table.

\begin{table*}[ht]
    \footnotesize
    \centering
    \setlength\tabcolsep{0pt} 
    \begin{tabularx}{\textwidth}{p{3.6cm}|*{20}{>{\centering\arraybackslash}X}|>{\centering\arraybackslash}X}
    \toprule
        Methods & \rotatebox{90}{aero} & \rotatebox{90}{bike} & \rotatebox{90}{bird} & \rotatebox{90}{boat} & \rotatebox{90}{bottle} & \rotatebox{90}{bus} & \rotatebox{90}{car} & \rotatebox{90}{cat} & \rotatebox{90}{chair} & \rotatebox{90}{cow} & \rotatebox{90}{table} & \rotatebox{90}{dog} & \rotatebox{90}{horse} & \rotatebox{90}{mbike} & \rotatebox{90}{person} & \rotatebox{90}{plant} & \rotatebox{90}{sheep} & \rotatebox{90}{sofa} & \rotatebox{90}{train} & \rotatebox{90}{tv} & \rotatebox{90}{mean} \\
    \midrule
        \multicolumn{22}{l}{VOC 2007 results:} \\
        OICR VGG16[31] & 81.7 & 80.4 & 48.7 & 49.5 & 32.8 & 81.7 & \textbf{85.4} & 40.1 & 40.6 & 79.5 & 35.7 & 33.7 & 60.5 & 88.8 & 21.8 & 57.9 & \textbf{76.3} & 59.9 & 75.3 & 81.4 & 60.6 \\
        PCL-OB-G VGG16[30] & 79.6 & \textbf{85.5} & 62.2 & 47.9 & 37.0 & \textbf{83.8} & 83.4 & 43.0 & 38.3 & \textbf{80.1} & 50.6 & 30.9 & 57.8 & \textbf{90.8} & 27.0 & \textbf{58.2} & 75.3 & 68.5 & \textbf{75.7} & 78.9 & 62.7 \\
        TS$^2$C[36] & \textbf{84.2} & 74.1 & 61.3 & \textbf{52.1} & 32.1 & 76.7 & 82.9 & 66.6 & 42.3 & 70.6 & 39.5 & 57.0 & 61.2 & 88.4 & 9.3 & 54.6 & 72.2 & 60.0 & 65.0 & 70.3 & 61.0 \\
    \hline
        OICR Ens.+FRCNN[31] & \textit{85.8} & 82.7 & 62.8 & 45.2 & 43.5 & \textit{84.8} & \textit{87.0} & 46.8 & 15.7 & \textit{82.2} & 51.0 & 45.6 & 83.7 & 91.2 & 22.2 & \textit{59.7} & 75.3 & 65.1 & 76.8 & 78.1 & 64.3 \\
        PCL-OB-G Ens.+FRCNN[30] & 83.8 & 85.1 & 65.5 & 43.1 & \textit{50.8} & 83.2 & 85.3 & 59.3 & 28.5 & \textit{82.2} & 57.4 & 50.7 & \textit{85.0} & \textit{92.0} & 27.9 & 54.2 & 72.2 & 65.9 & \textit{77.6} & 82.1 & 66.6 \\
    \hline
        \textbf{Ours} \rule{0pt}{1.01em} & 82.4 & 64.6 & \textbf{70.0} & 50.3 & \textbf{46.7} & 77.4 & 78.7 & \textbf{78.0} & \textbf{56.6} & 77.3 & \textbf{69.5} & \textbf{66.7} & \textbf{69.0} & 81.2 & \textbf{33.3} & 49.8 & 76.0 & \textbf{70.3} & 70.9 & \textbf{86.3}  & \textbf{67.8} \\
    \Xhline{2\arrayrulewidth}
        \multicolumn{21}{l}{VOC 2012 results:} & \rule{0pt}{1.1em} \\
        OICR VGG16[31] & 86.2 & \textbf{84.2} & 68.7 & \textbf{55.4} & 46.5 & 82.8 & 74.9 & 32.2 & 46.7 & \textbf{82.8} & 42.9 & 41.0 & 68.1 & 89.6 & 9.2 & 53.9 & 81.0 & 52.9 & 59.5 & 83.2 & 62.1 \\
        PCL-OB-G VGG16[30] & 77.2 & 83.0 & 62.1 & 55.0 & \textbf{49.3} & 83.0 & \textbf{75.8} & 37.7 & 43.2 & 81.6 & 46.8 & 42.9 & \textbf{73.3} & \textbf{90.3} & \textbf{21.4} & \textbf{56.7} & \textbf{84.4} & 55.0 & 62.9 & 82.5 & 63.2 \\
        TS$^2$C[36] & 79.1 & 83.9 & 64.6 & 50.6 & 37.8 & \textbf{87.4} & 74.0 & 74.1 & 40.4 & 80.6 & 42.6 & \textbf{53.6} & 66.5 & 88.8 & 18.8 & 54.9 & 80.4 & 60.4 & \textbf{70.7} & 79.3 & 64.4 \\
        \hline
        OICR Ens.+FRCNN[31] & \textit{89.3} & 86.3 & 75.2 & \textit{57.9} & 53.5 & 84.0 & \textit{79.5} & 35.2 & 47.2 & \textit{87.4} & 43.4 & 43.8 & 77.0 & 91.0 & 10.4 & 60.7 & \textit{86.8} & 55.7 & 62.0 & 84.7 & 65.6 \\
        PCL-OB-G Ens.+FRCNN[30] & 86.7 & \textit{86.7} & 74.8 & 56.8 & \textit{53.8} & 84.2 & 80.1 & 42.0 & 36.4 & 86.7 & 46.5 & \textit{54.1} & \textit{87.0} & \textit{92.7} & \textit{24.6} & \textit{62.0} & 86.2 & 63.2 & \textit{70.9} & 84.2 & \textit{68.0} \\
        \hline
        \textbf{Ours} \rule{0pt}{1.01em} & \textbf{87.9} & 70.1 & \textbf{76.6} & 54.7 & 48.9 & 80.8 & 72.8 & \textbf{76.5} & \textbf{51.9} & 69.6 & \textbf{64.7} & 49.8 & 63.7 & 83.1 & 21.1 & 55.2 & 80.9 & \textbf{75.5} & 62.3 & \textbf{85.9} & \textbf{66.6} \\
    \Xhline{2\arrayrulewidth}
    \end{tabularx}
    \caption{\textbf{Correct localization (in \%) on the Pascal VOC trainval set}. The top shows VOC 2007 and the bottom shows VOC 2012 results. The best single model is in \textbf{bold}, and best ensemble in \textit{italics}.} 
    \label{tab:result_voc07_corloc}
\end{table*}

\newpage
\section{Training with COCO captions: qualitative examples}

We provide more qualitative examples on the COCO \textit{val} set. We compare the \textsc{ExactMatch} and our \textsc{EM+TextClsf} (see paper Sec. 4.2 for details) in a side-by-side manner in the following figure. Qualitatively, our proposed method \textsc{EM+TextClsf} provides better detection results than the baseline \textsc{ExactMatch}.

\begin{figure}[ht]
    \centering
    \includegraphics[width=1.0\linewidth]{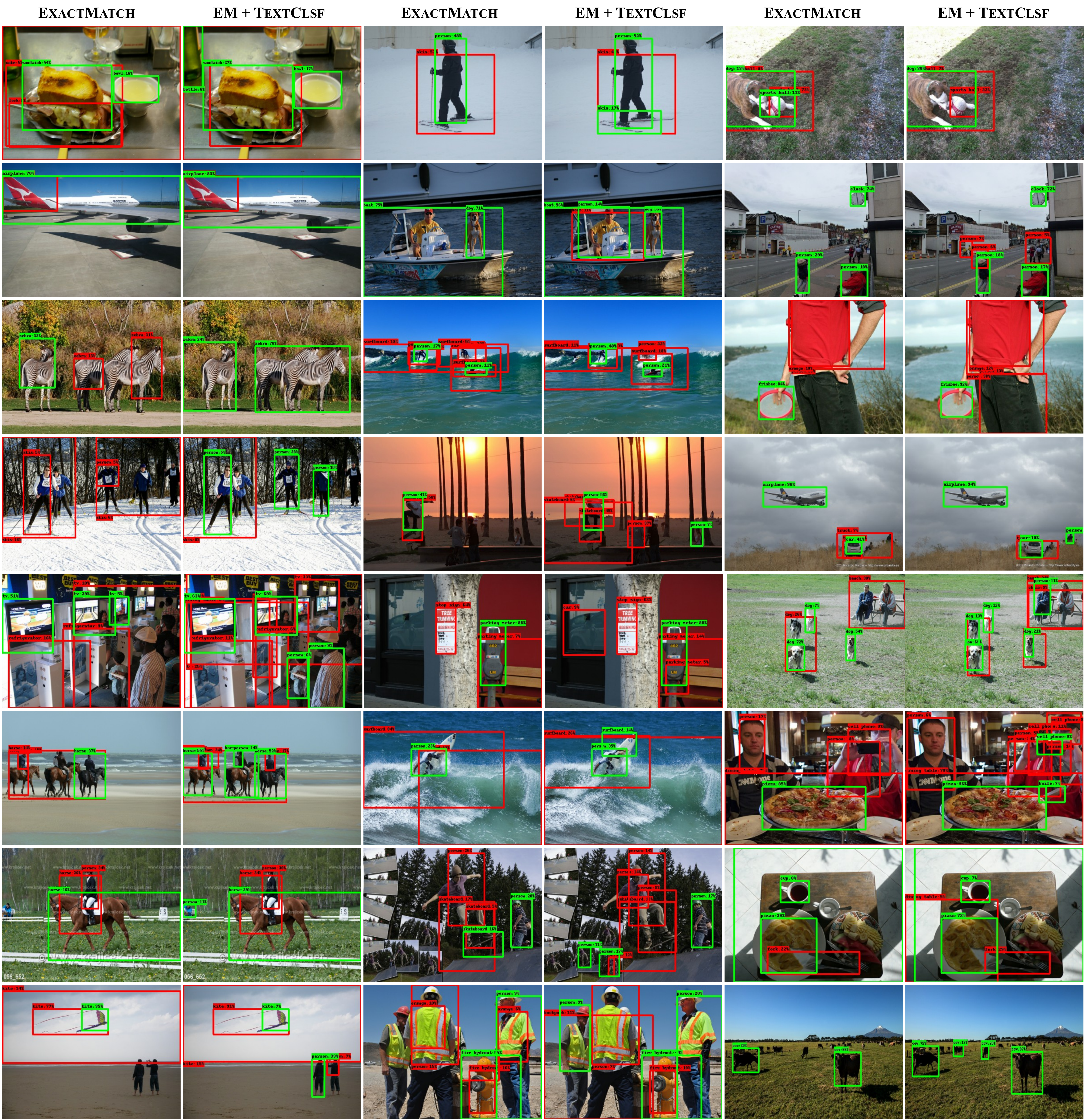}
    \caption{\textbf{Visualization of our Cap2Det model results}. We show boxes with confidence scores $>$ 5\%. Green boxes denote correct detection results ($IoU > 0.5$) while red boxes indicate incorrect ones.}
    \label{fig:coco_demo}
\end{figure}

\newpage
\section{Training with Flickr30K captions: qualitative examples}

We provide qualitative examples on the Flickr30K dataset in the following figure. We show the pseudo labels predicted by our label inference module. As a comparison, the \textsc{ExactMatch} fails to recall most of the image-level labels, while image-level supervisions are still accurate if we use the method in Sec. 3.1 (\textsc{EM+TextClsf}). Please note that there is NO image-level label on the Flickr30K dataset and our inference module purely transfers textual knowledge from the COCO, with NO training on Flickr30K.

\begin{figure}[ht]
    \centering
    \includegraphics[width=0.7\linewidth]{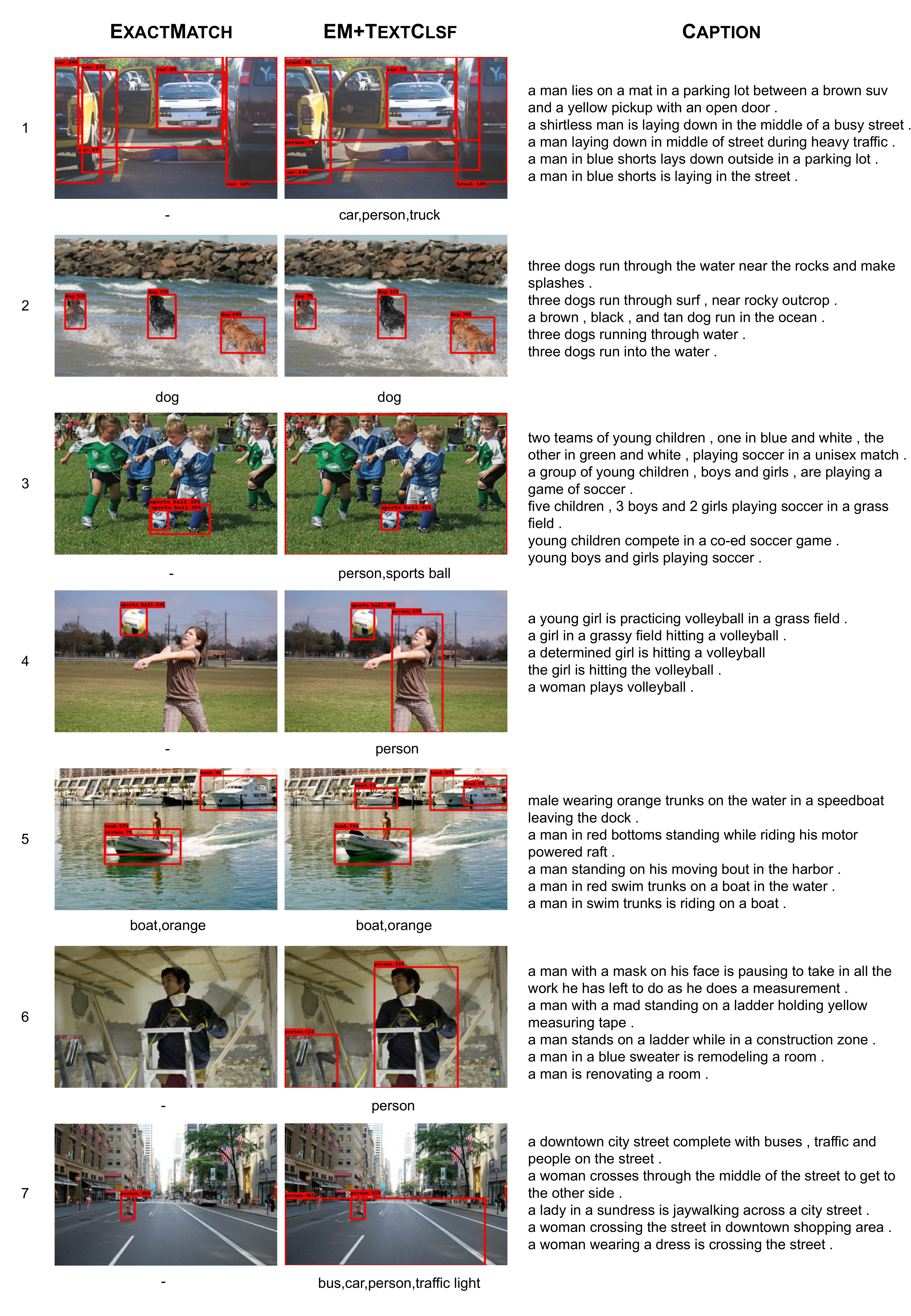}
    \caption{\textbf{Visualization of our Cap2Det model results}. We show boxes with confidence scores $>$ 5\%. We also show pseudo labels extracted from textual descriptions. Please note that there is neither instance-level nor image-level object labels in Flickr30K, but our label inference module fills in this gap.}
    \label{fig:flickr30k_demo}
\end{figure}

\end{document}